\pdfoutput=1

\documentclass[11pt]{article}

\usepackage[preprint]{coling}

\usepackage{times}
\usepackage{latexsym}

\usepackage[T1]{fontenc}

\usepackage[utf8]{inputenc}

\usepackage{microtype}

\usepackage{inconsolata}
\usepackage{graphicx}

\usepackage{algorithm}
\usepackage{algorithmic}
\usepackage{xcolor}
\usepackage{amsmath}
\usepackage{multirow}
\usepackage{array}
\usepackage{makecell}
\usepackage{float}
\usepackage{booktabs} 
\usepackage{rotating} 
\usepackage{tabularx} 
\usepackage{amsfonts}
\usepackage{subfigure}
\usepackage{comment}
\usepackage{enumitem}
\usepackage{ragged2e}
\usepackage{tikz}

\newcommand*\circled[1]{\tikz[baseline=(char.base)]{
            \node[shape=circle,draw,inner sep=2pt] (char) {#1};}}

\newcommand*{\note}[1]{\textcolor{red}{#1}}
\floatstyle{ruled}
\newfloat{listing}{tb}{lst}{}
\floatname{listing}{Listing}
\DeclareMathOperator*{\concat}{%
    \mathchoice%
        {\Big\Vert}%
        {\big\Vert}%
        {\Vert}%
        {\Vert}%
}
\title{GRL-Prompt: Towards Knowledge Graph based Prompt Optimization via Reinforcement Learning}

\author{Yuze Liu$^{1}$, Tingjie Liu$^{2}$, Tiehua Zhang$^{3*}$, Youhua Xia$^{2*}$, Jinze Wang$^{4}$,\\ \textbf{Zhishu Shen$^{5}$}, \textbf{Jiong Jin$^{4}$} \textbf{and Fei Richard Yu$^{6}$}
\\ \normalsize{$^{1}$Ant Group, Shanghai, China}
\\ \normalsize{$^{2}$Guangdong Laboratory of Artificial Intelligence (SZ), Shenzhen, China}
\\ \normalsize{$^{3}$School of Computer Science and Technology, Tongji University, Shanghai, China}
\\ \normalsize{$^{4}$Computing and Engineering Technologies, Swinburne University of Technology, Melbourne, Australia}
\\ \normalsize{$^{5}$School of Computer Science and Artificial Intelligence, Wuhan University of Technology, Wuhan, China} \\
\normalsize{$^{6}$College of Computer Science and Software Engineering, Shenzhen University, Shenzhen, China}
}

\begin{document}
\maketitle
\begin{abstract}
Large language models (LLMs) have demonstrated impressive success in a wide range of natural language processing (NLP) tasks due to their extensive general knowledge of the world. Recent works discovered that the performance of LLMs is heavily dependent on the input prompt. However, prompt engineering is usually done manually in a trial-and-error fashion, which can be labor-intensive and challenging in order to find the optimal prompts. To address these problems and unleash the utmost potential of LLMs, we propose a novel LLMs-agnostic framework for prompt optimization, namely GRL-Prompt, which aims to automatically construct optimal prompts via reinforcement learning (RL) in an end-to-end manner. To provide structured action/state representation for optimizing prompts, we construct a knowledge graph (KG) that better encodes the correlation between the user query and candidate in-context examples. Furthermore, a policy network is formulated to generate the optimal action by selecting a set of in-context examples in a rewardable order to construct the prompt. Additionally, the embedding-based reward shaping is utilized to stabilize the RL training process. The experimental results show that GRL-Prompt outperforms recent state-of-the-art methods, achieving an average increase of 0.10 in ROUGE-1, 0.07 in ROUGE-2, 0.07 in ROUGE-L, and 0.05 in BLEU.
\let\thefootnote\relax\footnotetext{$^{*}$ Corresponding authors.}

\end{abstract}
\vspace{-2mm}
\section{Introduction}
With ongoing advancements in large language models (LLMs), such as GPT-3~\cite{brown2020language} and LLaMA~\cite{touvron2023open}, optimizing query-guided prompts, especially in the realm of in-context learning (ICL), has been crucial in unlocking the full potential of LLMs, leading to significant improvements across various downstream natural language processing (NLP) tasks.~\cite{liu2021makes}. In the standard paradigm of ICL, a small collection of examples, each of which comprises small piece of contextual texts in natural language, is generally prepared to construct the prompt. This prompt effectively empowers LLMs to excel in tasks such as natural language understanding~\cite{ye2023compositional} and the logic reasoning~\cite{lu2022dynamic}. In contrast to many previous studies that concentrate on fine-tuning a large number of trainable parameters in the model for various downstream tasks~\cite{weifinetuned}, in-context learning (ICL) enables LLMs to achieve competitive performance without relying on gradient-based training, thus avoiding over-reliance on computational resources such as GPUs. An imperative issue in in-context learning is how to select the most suitable examples to construct prompts that effectively improve the LLM’s performance across various tasks~\cite{xu2024context}.

Previous research has addressed this challenge primarily through heuristic handcrafting, random selection, and retrieval-based selection methods. The heuristic handcrafting approach involves manually creating examples for a particular task, which can be both labor-intensive and time-consuming~\cite{wei2022chain}. Some studies have randomly select in-context examples from the training dataset to offer additional contextual information, with the goal of directing LLMs toward the desired responses~\cite{brown2020language}. Unfortunately, the randomness of the selection strategy introduces uncertainty and noise into the prompt, leading to a degradation in the performance of the LLMs. Retrieval-based selection attempts to tackle this issue by obtaining semantically similar examples using static methods such as K-nearest neighbors~\cite{nie2022improving}. However, this method fails to thoroughly explore the prompt space, leading to limited effectiveness in complex tasks. It has been pointed out in recent studies that retrieval-based selection methods face the challenge of overlooking the permutations of in-context examples, which can lead to a degradation in the performance of LLMs ~\cite{lu2022fantastically,xu2024context}.

To alleviate this challenge, we propose a novel LLMs-agnostic framework for prompt optimization, namely GRL-Prompt, which aims to automatically construct optimal prompts via reinforcement learning (RL) in an end-to-end manner. Our GRL-Prompt framework comprises two key components: 1) a knowledge graph built from the user instruction and candidate in-context examples, utilizing a heterogeneous graph neural network to encode the structured embedding representations of the graph; and 2) a policy network that includes a pairwise edge classifier and an in-context matching network, generating an optimal sequence of in-context examples to create the most effective prompts based on the constructed knowledge graph. The knowledge graph and the policy network are updated collaboratively in an end-to-end manner. Our contributions can be summarized as follows:
\begin{itemize}
    \vspace{-2mm}
    \item We propose a novel LLMs-agnostic framework GRL-Prompt for prompt optimization. The knowledge graph is initially constructed to effectively capture the correlation between the user instruction and candidate in-context examples through heterogeneous graph learning. The designated policy network allows the GRL-Prompt to function as an agent that interacts iteratively with the LLMs.
    \vspace{-2mm}
    \item We formulate a policy network that incorporates a pairwise edge classifier (PEC) and an in-context matching network (ICMN) to dynamically learn the policy for generating an order-sensitive sequence of in-context samples. PEC is designed to classify the order of examples, while ICMN aims to select the relevant examples to construct the optimal prompt based on the structured representation of a constructed knowledge graph. Moreover, we design the embedding-based reward shaping to stabilize the RL training process.
    \vspace{-2mm}
    \item We conduct extensive experiments on two distinct datasets, demonstrating that GRL-Prompt outperforms state-of-the-art baselines in in-context learning. Furthermore, we have made our source code publicly available to contribute further to advancements in this field (find the source code as supplementary materials).
\end{itemize}

\section{Related Work}
\vspace{-3mm}
In recent years, LLMs have achieved remarkable progress in the field of NLP~\cite{lewis2019bart, raffel2020exploring}. Notably, GPT-3~\cite{brown2020language} has demonstrated exceptional capabilities in few-shot ICL~\cite{dong2022survey}, where it learns tasks with only a few examples as demonstrations, without requiring any gradient updates~\cite{liu2021makes, min2022rethinking}.

To further unleash the potential of LLMs without updating the large amount of parameters in the model, a series of studies have focused on developing prompt optimization techniques to guide models toward generating more accurate and relevant outputs with high-quality input prompts.~\citet{wei2022chain} investigated how LLMs handle few-shot prompting for reasoning tasks using prompts consisting of triples—input, chain of thought, and output—and examined chain-of-thought prompting as a straightforward and widely applicable technique for improving reasoning in these models.
Following that,~\citet{wang2023boosting} proposed a chain-of-knowledge prompting approach that decomposes reasoning chains generated by LLMs into several evidence triples and explanatory hints, thereby enhancing the reasoning capabilities. \citet{li2024unlocking} developed a tabular prompting method called TableIE, which reframes the relational triple extraction task as a table generation problem and has demonstrated promising results in ICL. These methods have demonstrated how prompt optimization enhances the logical reasoning and content generation capabilities of LLMs.

Despite these impressive successes, the aforementioned works still face challenges when tackling tasks that require deep and complex reasoning~\cite{xu2024context}. Knowledge Graphs (KGs)~\cite{chen2020review} serve as an ideal foundation for prompt optimization by providing structured knowledge representation. KGs encapsulate domain knowledge through entities and relationships, providing LLMs with rich contextual information and reasoning pathways. However, relying on knowledge graphs alone for prompt construction still has its limitations, as knowledge graph-based prompts often overlook the context-dependence of natural language, resulting in degraded performance on complex tasks. Therefore, it is essential to optimize knowledge graph-based prompts through interactive feedback from LLMs~\cite{li2024enhanced}.

To address these challenges, several studies have started to explore reinforcement learning for prompt optimization.~\citet{deng2022rlprompt} proposes RLPROMPT, a discrete prompt optimization approach with RL. This method formulates a parameter-efficient policy network that generates the optimized discrete prompt by training with the designated reward function.~\citet{qi2023pillow} introduces PILLOW, a prompt-matching framework enhanced by RL. This framework utilizes a matching network to select prompts from a user-defined pool, concatenates them with user instructions as input, and performs inference with LLMs to improve the efficiency of instruction fine-tuning. Similiarly, PROMPTPG employs a RL-based policy gradient to learn how to select in-context examples from a limited amount of training data and then dynamically constructs the corresponding prompt~\cite{lu2022dynamic}. 

\noindent\textbf{Relation to Existing ICL methods}\quad
Compared to the random selection in the ICL paradigm~\cite{brown2020language,wei2022chain}, GRL-Prompt updates the policy network iteratively to generate an optimal sequence of in-context examples for prompt optimization. ICL-kNN~\cite{nie2022improving}, on the other hand, uses a static clustering method (k-nearest neighbor) to select in-context examples, which also lacks the dynamic learning on the policy network. This method thus fails to thoroughly explore the prompt space, leading to limited effectiveness in complex tasks. Another issue is that the order of the in-context samples introduces variation in the performance of LLMs. GRL-Prompt can dynamically learn the policy for generating an order-sensitive sequence of in-context samples because its policy network has two key components: a PEC and an ICMN. In contrast, PromptPG~\cite{lu2022dynamic} and Pillow~\cite{qi2023pillow} only select a fixed-length set of in-context samples. Moreover, the constructed knowledge graph, which encodes the correlation between the user query and the candidate examples, provides a structured representation for GRL-Prompt to optimize prompts.

\begin{figure*}[tb!]
    \vspace{-6mm}
    \centering
    \includegraphics[width=0.9\linewidth]{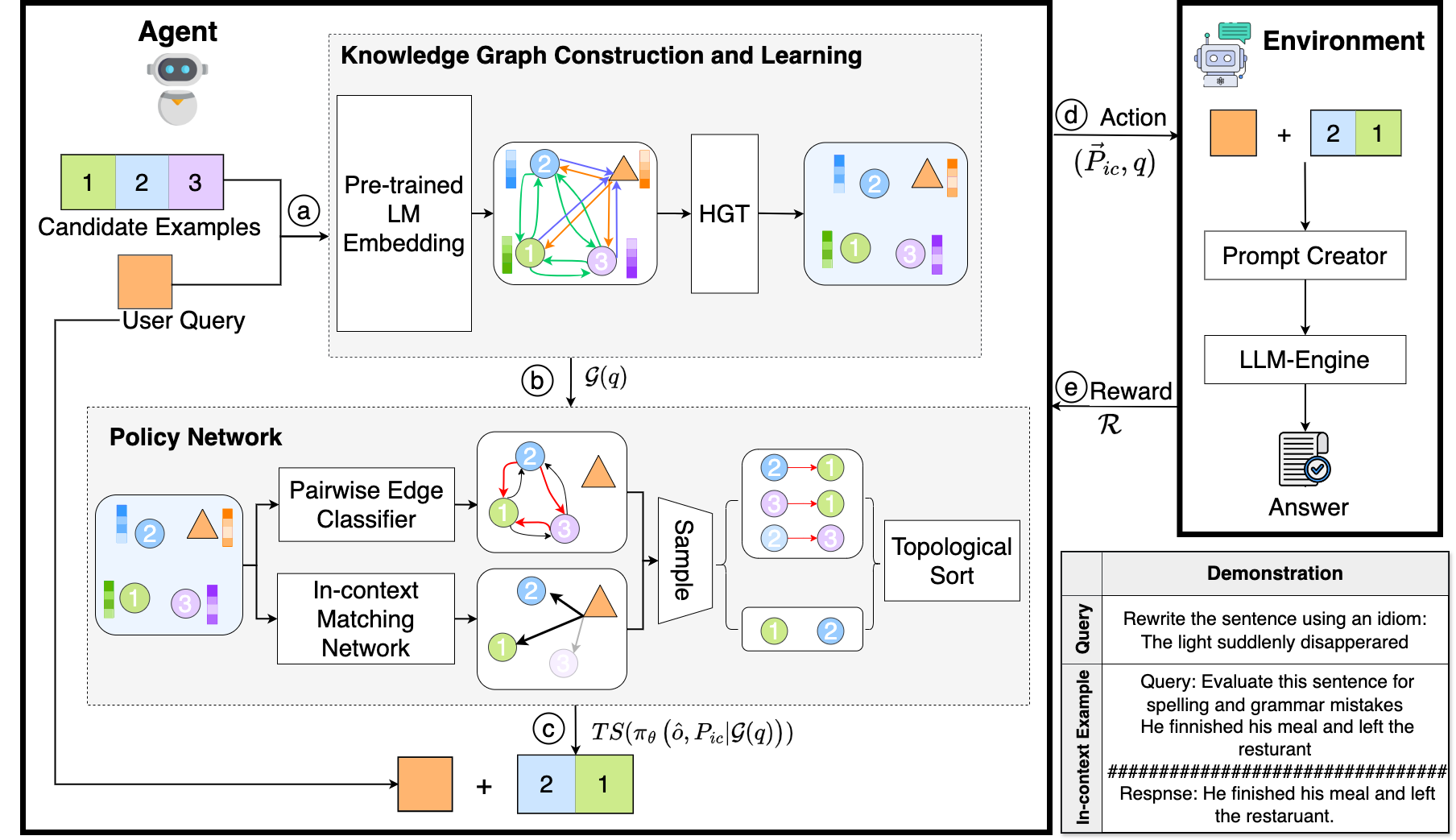}
  \caption{An overview of GRL-Prompt. The circle and triangle represent the candidate nodes and the query node, respectively. The green arrow lines denote candidate-to-candidate edges, the orange arrow lines indicate query-to-candidate edges, and the blue arrow lines represent candidate-to-query edges in the knowledge graph.}
  \vspace{-2mm}

  \label{fig:all} 
\end{figure*} 
\vspace{-2mm}
\section{Preliminary and Problem Definition}
In this section, we first introduce the relevant  definitions and then formulate the key problems
associated with prompt optimization.

\noindent\textbf{Knowledge Graph}\quad
 A knowledge graph is defined as $\mathcal{G} = (\mathcal{V},\mathcal{E},\mathcal{R})$, with nodes $v_{i}\in\mathcal{V}$ and directed edges $(v_{i}, r_{ij}, v_{j})\in\mathcal{E}$, where $r_{ij} \in\mathcal{R}$ is the relation type of the edge between $v_i$ and $v_j$.

\noindent\textbf{Reinforcement Learning}\quad
The RL is formulated as a Markov Decision Process (MDP), which is a sequential decision-making mathematical model in which actions affect the current short-term rewards, subsequent states, and future rewards~\cite{sutton1998introduction}. The MDP is defined as the tuple $\{\mathcal{S}, \mathcal{A}, \mathcal{T}, \mathcal{R}, \gamma\}$, where $s\in\mathcal{S}$ and $a\in\mathcal{A}$ denote state and action, respectively. $\mathcal{T}:\mathcal{S}\times\mathcal{A}\rightarrow p(\mathcal{S})$ is identified by the states transition. $p(\cdot)$ denotes the distribution of the states. $\mathcal{R}$ is the reward function, with $\gamma\in[0, 1]$ as the discount factor.

\noindent\textbf{In-context Learning}\quad
ICL is a paradigm that facilitates language models to learn tasks using only a few in-context examples presented in the form of natural language~\cite{brown2020language}. Formally, the sequence of in-context examples is denoted as $\vec{\emph{P}}_{ic} = [p_{ic}^{1}, ..., p_{ic}^{K}]$, where $K$ is the number of in-context examples provided to construct the prompt.

\noindent\textbf{Prompt Optimization Problem}\quad
Given a user query text $q\in\emph{Q}$ and a set of candidate examples $\emph{P}_{cand}$, we aim to learn a prompt optimization strategy $f_{opt} (q, \emph{P}_{cand})\rightarrow \emph{P}_{ic}$. Each candidate example $p_{cand}\in\emph{P}_{c}$ is a triple of $(query, context, response)$,  where $query$ denotes the user query, $context$ represents the extra information provided by the users (optional), and $response$ is the expected answer. The generated sequence of in-context examples provided for LLMs can be used for a variety of downstream NLP tasks.

\section{Methodology}
In this section, we introduce GRL-Prompt, an LLMs-agnostic framework for prompt optimization via RL. As illustrated in Figure~\ref{fig:all}, a knowledge graph is constructed using candidate examples and a user query, from which a heterogeneous graph neural network is employed to encode their correlations into high-dimensional embedding representations for the downstream component. The policy network in the agent learns to find the optimal sequence of in-context examples from the constructed knowledge graph, aiming to maximizing the prediction rewards for the user query when interacting with LLM environment. Additionally, an embedding-based reward shaping is utilized to stabilize the RL training process.
\subsection{Knowledge Graph for Prompt Construction}
The correlations between in-context examples and the user query influence the performance of LLMs in the ICL paradigm \cite{nie2022improving}. GRL-Prompt explores the application of KG to better encode the correlations between candidate examples and the user query, facilitating prompt construction. The user query and candidate examples are provided to the KG construction and learning module, as indicated by \circled{a} in  Figure~\ref{fig:all}.
\subsubsection{Knowledge Graph Construction}
Since there is no concept of nodes in the set of candidate examples $\emph{P}_{c}$ and the user query $q$, we treat each candidate example $p_{c}^{i}\in\emph{P}_{cand}, i\in[1,..., N]$ as a candidate node $v_{c}^{i}$ and the user query $q$ as query node $v_{q}$. The candidate node $v_{c}^{i}$ and the query node $v_{q}$ are represented as a circle and a triangle in Figure~\ref{fig:all}, respectively. The number of the candidate examples is $N$. The node set $\mathcal{V} = \{v_{c}^{i}\}_{i = 1}^{N}\cup v_q$ in the knowledge graph $\mathcal{G}(q)$ consists of two types of nodes: candidate node $v_{c}$ and query node $v_{q}$. We utilize a pre-trained language model, such as BERT~\cite{lu2022dynamic} and RoBERTa~\cite{ghosal2021stack}, to generate the initial node embeddings in the knowledge graph, which is defined as:
\begin{equation}
    \emph{X}^{0} = PreLM(\mathcal{V})
\end{equation}
Here, we use the values from the final layer of the pre-trained language model as the initial node embeddings $\emph{X}^{0}\in\mathcal{R}^{(N+1)\times d}$.

We construct three types of edges: candidate-to-candidate edges, query-to-candidate edges and candidate-to-query edges, which are based on the constituent nodes to encode different relations. Each candidate node $v_{c}^{i}\in\mathcal{V}$ is connected to all other candidate nodes $v_{c}^{j} ( j\neq i$) with relation $r_{cc}$. The directed candidate-to-candidate edge is denoted as $(v_{c}^{i}, r_{cc}, v_{c}^{j})$, represented by the green arrow lines in Figure 1. Our formulation results in bidirectional edges between each pair of candidate nodes, i.e. both $(v_{c}^{i}, r_{cc}, v_c^{j})$ and $(v_{c}^{j}, r_{cc}, v_c^{i})\in\mathcal{E}$. The query-to-candidate edge is constituted by the query node $v_q$ and the candidate node $v_c$, which is denoted as $(v_q, r_{qc}, v_c)$. Meanwhile, each candidate node $v_{c}^{i}\in\mathcal{V}$ is bidirectionally connected to the query node $v_q$, leading to the candidate-to-query edges $(v_{c}^{i}, r_{cq}, v_q)$ and query-to-candidate edges $(v_q, r_{qc}, v_c^{i})$, respectively. These two types of edges are represented by the blue arrow lines and the orange arrow lines, respectively.
\subsubsection{Knowledge Graph Learning}
To capture the complex interactions between the candidate examples and the user query in the constructed knowledge graph, we utilize a two-layer Heterogeneous Graph Transformer (HGT)~\cite{hu2020heterogeneous} to perform knowledge graph learning. The details of HGT can be found Appendix~\ref{adx:a_1}.

\subsection{RL-based Prompt Optimization}

Recent works have unveiled the high sensitivity of LLMs to prompts~\cite{li2024dialoguepromptingpolicygradientbaseddiscrete,bach2022promptsourceintegrateddevelopmentenvironment}. Recent research has shown that the performance of LLMs with in-context learning can be highly unstable across different selections of in-context examples and permutations of those examples~\cite{liu2021makes,lu2022fantastically}. Apart from that, finding the optimal prompt for various LLMs in different tasks is usually done manually in a trial-and-error fashion, which can be labor-intensive and challenging. To alleviate this issue, we design a policy network that utilizes the structural correlations in the constructed knowledge graph, as indicated by \circled{b} in Figure~\ref{fig:all}, to optimize prompts within the RL paradigm, avoiding brute-force searching or manually designed heuristics.
\subsubsection{Policy Network}
We design a policy network $\pi_{\theta}\left(\hat{o}, \emph{P}_{ic}|\mathcal{G}(q)\right)$ that incorporates a PEC to predict the relative order of any two candidate examples, utilizing the structural representation of the candidate nodes within the knowledge graph. Furthermore, an ICMN in policy network assesses the probability of each candidate example being selected as an in-context example within the constructed knowledge graph.

PEC uses the node embedding in the constructed knowledge graph to classify the order between two corresponding candidate examples. For instance, taking $p_c^{i}$ and $p_c^{j}$ as two candidate examples, where $i\neq j$, and their corresponding nodes in the constructed knowledge graph $\mathcal{G}$ are $v_c^{i}$ and $v_c^{j}$, respectively. In this formulation, PEC considers the bidirectional edges between $v_c^i$ and $v_c^j$ in $\mathcal{E}$ of $\mathcal{G}(q)$ --- $(v_c^{i}, r_{ij}, v_{c}^j)$ and $(v_c^{j}, r_{ji}, v_{c}^i)$. The classification objective is then to compare the pair scores of two edges that are reversed in direction, which is denoted as:
\begin{equation}
    f_{pec}(v_c^{i}, v_c^{j})=\max(\left[ps(v_c^{i}, v_{c}^{j}),ps(v_c^{j}, v_{c}^{i})\right])
\vspace{-2mm}
\end{equation}

If $ps(v_c^{i}, v_{c}^{j})>ps(v_c^{j}, v_{c}^{i})$, then we predict candidate example $p_c^{i}$ appears earlier than $p_c^{j}$ $(p_c^{i}\rightarrow p_c^{j})$, or vice versa  $(p_c^{j}\rightarrow p_c^{i})$. Naturally, the pair score function $ps(\cdot, \cdot)$ must be sensitive to the order of its inputs, and the output represents the probability of the order between two candidate examples. We define the $ps$ function as follows:
    \begin{equation}
    ps(v_c^{i},v_{c}^{j})\!=\!\frac{e^{\sin\!{(\emph{X}(v_c^{i})-\emph{X}(v_c^{j}))}\cdot\emph{w}}}{e^{\sin\!{(\emph{X}(v_c^{i})-\emph{X}(v_c^{j}))}\cdot\emph{w}}+\!e^{\sin\!{(\emph{X}(v_c^{j})-\emph{X}(v_c^{i}))}\cdot\emph{w}}}
\end{equation}
where $\emph{w}\in\mathcal{R}^d$ is the learnable parameter of the function. 
\begin{table*}[ht]
\vspace{-6mm}
\centering
\scalebox{0.71}{
\begin{tabular}{llccclc|ccclc}
\hline
                        & \multicolumn{1}{c}{}                         & \multicolumn{5}{c|}{Alpaca}                                                                          & \multicolumn{5}{c}{Dolly}                                                     \\ \cline{3-12} 
\multirow{-2}{*}{Model} & \multicolumn{1}{c}{\multirow{-2}{*}{Method}} & ROUGE-1                             & ROUGE-2      & \multicolumn{2}{c}{ROUGE-L}      & BLEU         & ROUGE-1      & ROUGE-2      & \multicolumn{2}{c}{ROUGE-L}      & BLEU         \\ \hline
                        & Random-1                                     & 0.36                                & 0.17         & \multicolumn{2}{c}{0.29}         & 0.11         & 0.31         & 0.13         & \multicolumn{2}{c}{0.22}         & 0.05         \\
                        & Random-2                                     & 0.35                                & 0.17         & \multicolumn{2}{c}{0.27}         & 0.10         & 0.35         & 0.15         & \multicolumn{2}{c}{0.25}         & 0.07         \\
                        & PromptPG                                     & {0.38}                          & {0.18}   & \multicolumn{2}{c}{0.30}   & {0.11}   & {\underline{0.42}}   & {0.21}   & \multicolumn{2}{c}{{\underline{0.32}}}   & {0.11}   \\
                        & CoT                                          & 0.18                                & 0.07         & \multicolumn{2}{c}{0.13}         & 0.03         & 0.18         & 0.06         & \multicolumn{2}{c}{0.12}         & 0.02         \\& Pillow                                       & {0.40}                               & {0.19}         & \multicolumn{2}{c}{0.32}         & {0.13}         & 0.41         & 0.20         & \multicolumn{2}{c}{0.31}         & 0.11         \\
                        & UDR                                     & {\underline{0.42}}                               & {\underline{0.20}}        & \multicolumn{2}{c}{{\underline{0.33}}}         & {\underline{0.13}}         & 0.41         & {\underline{0.21}}       & \multicolumn{2}{c}{0.31}         & {\underline{0.11}}      \\ \cline{2-12}  \cline{2-12} 

\multirow{-8}{*}{GPT-4} & \textbf{Ours}                                & {0.44 \note{(+0.02)}} & 0.23 \note{(+0.03)} & \multicolumn{2}{c}{0.34 \note{(+0.01)}} & 0.15 \note{(+0.02)} & 0.43 \note{(+0.01)} & 0.21 \note{(+0.00)} & \multicolumn{2}{c}{0.32 \note{(+0.00)}} & 0.11 \note{(+0.00)} \\ \hline
                        & Random-1                                     & 0.41                                & 0.20         & \multicolumn{2}{c}{0.32}         & 0.13         & 0.38         & 0.18         & \multicolumn{2}{c}{0.30}         & 0.10         \\
                        & Random-2                                     & 0.44                                & 0.22         & \multicolumn{2}{c}{0.35}         & 0.15         & 0.38         & 0.17         & \multicolumn{2}{c}{0.29}         & 0.10         \\
                        & PromptPG                                     & {0.46}                          & {0.23}   & \multicolumn{2}{c}{0.37}   & {0.16}   & {\underline{ 0.45}}   & {0.24}   & \multicolumn{2}{c}{0.35}   & {\underline{0.14}}   \\
                        & CoT                                          & 0.35                                & 0.16         & \multicolumn{2}{c}{0.27}         & 0.09         & 0.33         & 0.14         & \multicolumn{2}{c}{0.24}         & 0.07         \\
                        & Pillow                                       & {\underline{0.47}}                                & 0.23         & \multicolumn{2}{c}{{\underline{0.38}}}         & 0.16         & 0.43         & 0.23         & \multicolumn{2}{c}{0.35}         & 0.13         \\
                        & UDR                                     & 0.46                               & {\underline{0.23}}         & \multicolumn{2}{c}{0.37}         & {\underline{0.16}}         & 0.43         & {\underline{0.24}}        & \multicolumn{2}{c}{{{\underline{0.35}}}}         & 0.13         \\ \cline{2-12} \cline{2-12} 
\multirow{-8}{*}{GPT-3} & \textbf{Ours}                                & 0.50 \note{(+0.03)}                        & 0.27 \note{(+0.04)} & \multicolumn{2}{c}{0.39 \note{(+0.01)}} & 0.19 \note{(+0.03)} & 0.47 \note{(+0.02)} & 0.26 \note{(+0.02)} & \multicolumn{2}{c}{0.37 \note{(+0.02)}} & 0.15 \note{(+0.01)} \\ \hline
                        & Random-1                                     & {0.32}                          & 0.14         & \multicolumn{2}{c}{0.25}   & {0.09}   & 0.21         & 0.08         & \multicolumn{2}{c}{0.16}         & 0.04         \\
                        & Random-2                                     & 0.30                                & 0.14         & \multicolumn{2}{c}{0.23}         & 0.08         & 0.26         & 0.09         & \multicolumn{2}{c}{0.18}         & 0.04         \\
                        & PromptPG                                     & 0.31                                & {0.14}   & \multicolumn{2}{c}{0.23}         & 0.08         & {0.38}   & {0.19}   & \multicolumn{2}{c}{0.30}  & {0.10}   \\
                        & CoT                                          & 0.16                                & 0.07         & \multicolumn{2}{c}{0.12}         & 0.03         & 0.16         & 0.06         & \multicolumn{2}{c}{0.12}         & 0.02         \\
                        &Pillow                                       & {0.33}                                & {0.16}         & \multicolumn{2}{c}{0.27}         & {0.11}         & 0.36         & 0.19         & \multicolumn{2}{c}{0.28}         & 0.10         \\
                        & UDR                                     & {\underline{0.41}}                                 & {\underline{0.20}}          & \multicolumn{2}{c}{\underline{0.33}}          & {\underline{0.13}}          & \underline{0.39}        & {\underline{0.19}}         & \multicolumn{2}{c}{{{\underline{0.30}}}}         & \underline{0.11}         \\ \cline{2-12} 
                        \cline{2-12} 
\multirow{-8}{*}{LLaMA} & \textbf{Ours}                                & 0.45 \note{(+0.04)}                        & 0.23 \note{(+0.03)} & \multicolumn{2}{c}{0.35 \note{(+0.02)}} & 0.15 \note{(+0.02)} & 0.41 \note{(+0.02)} & 0.21 \note{(+0.02)} & \multicolumn{2}{c}{0.31 \note{(+0.01)}} & 0.12 \note{(+0.01)} \\ \hline
\end{tabular}
}  
\caption{Results on GRL-Prompt on Alpaca and Dolly. The score differences that indicate better performance than the best baselines are marked with red color. Underlining indicates the value that achieves the best performance among all baselines}
\vspace{-4mm}
\label{tab:results}
\end{table*}

ICMN calculates probability of selecting each candidate examples $p_c^{i}\in P_{cand}$ for the set of in-context samples $P_{ic}$ based on the node embedding of the knowledge graph. For each candidate examples $p_c^{i}$, the corresponding node in the knowledge graph is $v_{c}^{i}$, and the user query $q$ corresponds to the query node $v_q$. The calculation process of the selecting probability is defined as follows:
\begin{equation}
    f_{icmn}(v_q,\!v_c^i)\!=\!sigmoid(\frac{\emph{X}(\!v_q\!)\!\cdot\!\emph{W}_{m}\!\cdot\!\emph{X}^{T}\!(v_c^i)\!}{\sqrt{d}})
    \label{eq:6}
\end{equation}
$\emph{W}_{m}\in\mathcal{R}^{d\times d}$ denotes the learnable matrix, and $sigmoid$ function maps the similarity between the candidate example and the user query into probability space. $\sqrt{d}$ acts as a scaling factor.

The policy network obtains the probability distribution for the generation of the sequence of in-context examples from the candidate examples set $P_{cand}$ via the component of ICMN and PEC, which is denoted as:
\begin{equation}
    p(\pi)\!=\!(\!\!\!\prod_{i\in[1,N]\atop i<j}\!\!\!\!f_{pec}(v_c^i, v_c^j))\!\!\times\!\!(\!\!\!\prod_{i\in[1,N]}\!\!\!\!f_{icmn}(v_q,\!v_c^i))\!\!
    \vspace{-3mm}
\end{equation}
where $N$ is the number of the set of candidate examples. Finally, given a user query $q$, the sequence of in-context examples is generated from the set of candidate examples according to the policy network
\begin{equation}
\vec{\emph{P}}_{ic}\sim TS(\pi_{\theta}\left(\hat{o}, \emph{P}_{ic}|\mathcal{G}(q)\right)), \vec{\emph{P}}_{ic}\in\mathcal{F}(\emph{P}_{cand})
\end{equation}
where $\mathcal{F}(\cdot)$ is the operator that constructs a set of all possible order-sensitive subsets from a given set. The total number of possible order-sensitive subsets from a set with $n$ elements is $\sum_{i = 1}^{n}\frac{n!}{(n-i)!}$~\cite{velleman1995permutations}. We use $\mathcal{F}(\emph{P}_{cand})$ to denote the state space in the RL in our formulation. $TS(\cdot)$ represents the topological sorting method~\cite{bommasani2020intrinsic}, which is used to obtain the final ordered sequence from the all the pairwise edges orders $\hat{o}$. If the PEC predicts that the order between the candidate example $p_c^i$ and $p_c^j$ is $(p_c^{i}\rightarrow p_c^{j})$, $TS(\cdot)$ ensures $p_c^i$ comes before $p_c^j$ in the final ordering. We sample the set of in-context examples $\emph{P}_{ic}$ and pairwise orders $\hat{o}$ associated with the in-context examples from the policy network, and the sequence of in-context examples $\vec{\emph{P}}_{ic}$ is generated by $TS(\cdot)$. 
\subsubsection{Reward Design}
Since we test GRL-Prompt on general text-to-text generation tasks, the reward is designed based on the evaluation of the responses from the LLMs using the in-context examples that we predict. The variation in the responses from black-box LLMs poses challenges to the training efficiency and convergence rate of the RL training process. To address these issues, we develop embedding-based reward shaping to smooth the reward function and stabilize the RL training, which can be expressed as:
\vspace{-1mm}
\begin{equation}
    \mathcal{R}(a, \hat{a})=\lambda\mathcal{R}_{m}(a,\hat{a})+(1-\lambda)\mathcal{R}_{e}(a,\hat{a})
\end{equation}
$\mathcal{R}_{m}$ represents fuzzy textual similarity, and $\mathcal{R}_{e}$ denotes the cosine embedding similarity based on sentence representations. $a$ represent the expected output. The generated response of LLMs denotes $\hat{a}$, which uses the predicted sequence of in-context examples $\vec{\emph{P}}_{ic}$ and the given user query $q$ to construct the prompt. $\lambda$ is the hyperparameter for adjusting the weight of two components in the reward function.

\begin{figure}[tb!]
\vspace{-4mm}
    \centering
    \scalebox{0.92}{
    \subfigure[\label{fig:2}
	Training loss on Alpaca.]
	{\includegraphics[width=1.75in,height=1.35in]{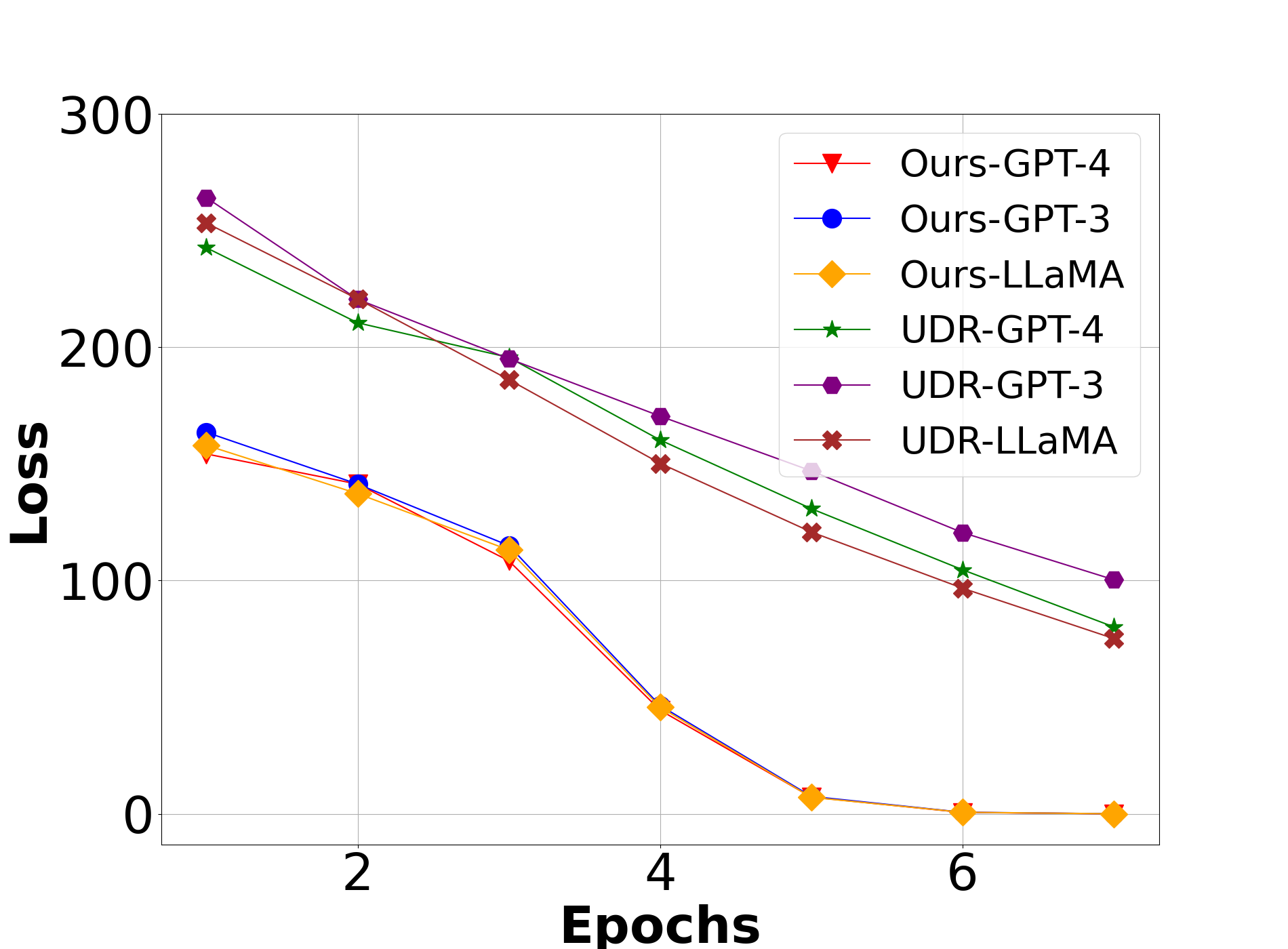}}
	\subfigure[\label{fig:3}
	Training loss on Dolly.] 
	{\includegraphics[width=1.75in,height=1.35in]{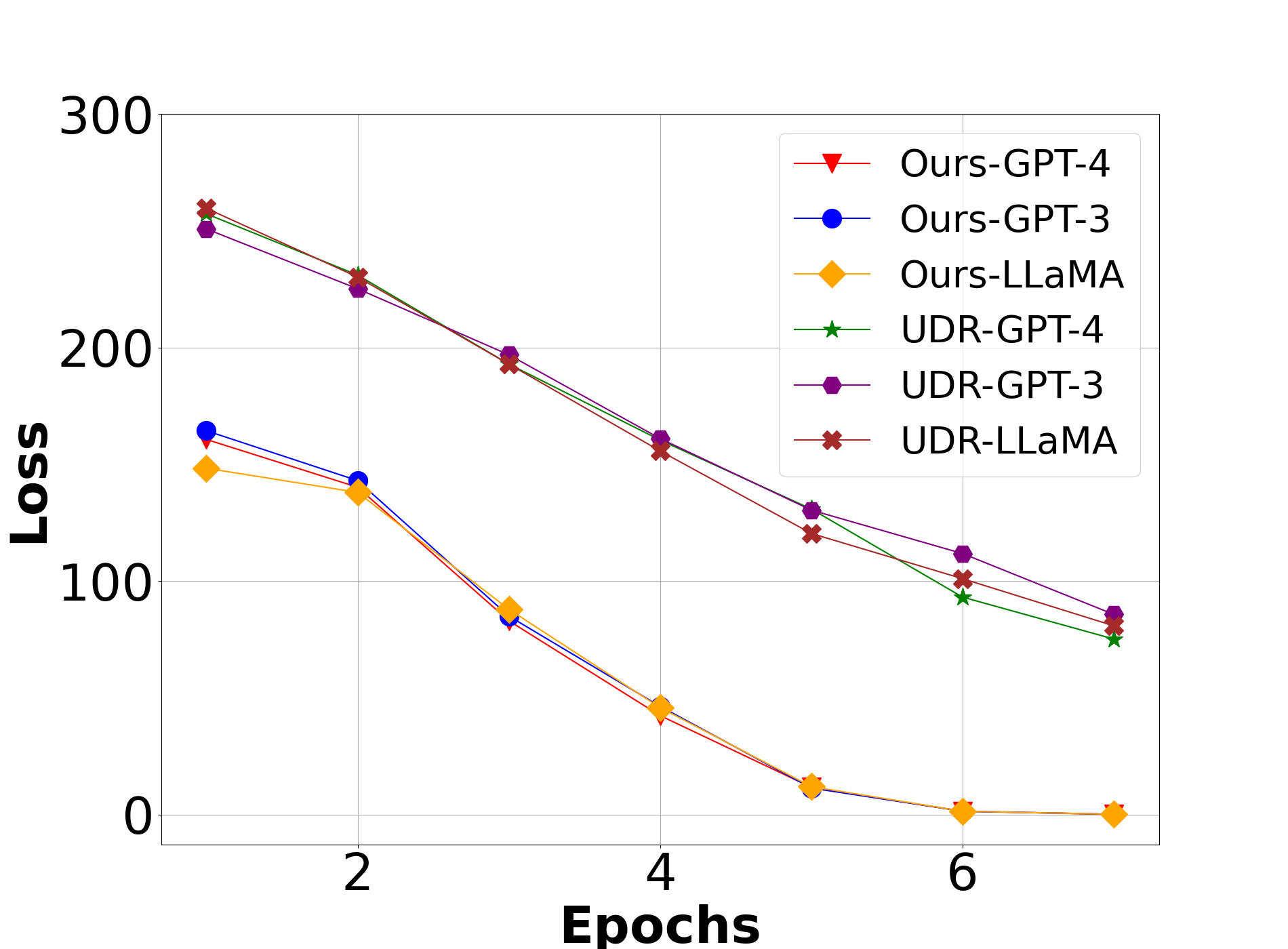}}
 }
     \vspace{-2mm}
 \caption{Results of training loss on different datasets.}
 \vspace{-2mm}
\end{figure}



\subsubsection{Training Process}
Given the set of training user query $\emph{Q}_{tr} = \{q^{i}\}_{i = 1}^{M}$ and their corresponding the expected responses $\{a^{1}, ..., a^{M}\}$, our goal is to maximize the expected reward of the generated response using the policy network $\mathbb{E}_{\vec{\emph{P}}_{ic}^{i}\sim\pi_{\theta}(\hat{o}^{i}, \emph{P}_{ic}^{i}|\mathcal{G}(q^{i}))}[\mathcal{R}(a^{i}, \hat{a}^{i})]$. We optimize the reward with respect to the parameters of the policy network using the policy gradient algorithm~\cite{williams1992simple}, which is defined as follows:
\vspace{-5mm}
\begin{equation}
\begin{split}
        &\nabla\mathbb{E}_{\vec{\emph{P}}_{ic}^{i}\sim\pi_{\theta}(\hat{o}^{i},\emph{P}_{ic}^{i}|\mathcal{G}(q^{i}))}[\mathcal{R}(a^{i},\hat{a}^{i})] \\
        &\!=\!\mathbb{E}_{\vec{\emph{P}}_{ic}^{i}\sim\pi_{\theta}(\hat{o}^{i}\!,\emph{P}_{ic}^{i}|\mathcal{G}(q^{i})\!)\!}\!\nabla\!_{\theta}\!\log(\!\pi_{\theta}(\hat{o}^{i},\!\emph{P}_{ic}^{i}|\mathcal{G}(q^{i})))\mathcal{R}(\!a^{i},\!\hat{a}^{i}\!)\\
        &\!\approx\!\!\frac{1}{m}\!\!\sum^{m}_{i=1}\!\nabla\!_{\theta}\!\log(\pi_{\theta}\!(\hat{o}^{i}\!,\!\emph{P}_{ic}^{i}|\mathcal{G}(q^{i})))\mathcal{R}(\!a^{i}\!,\!\hat{a}^{i}\!),\vec{\emph{P}}_{ic}^{i}\!\sim\! TS(\!\pi_{\theta}\!)\!
        \\[-4mm]
\end{split}
\vspace{-5mm}
\end{equation}
Here, $m$ denotes the size of each batch from our training set of user query $\emph{Q}_{tr}$. The learnable parameters in the knowledge graph and the policy network are updated iteratively as GRL-Prompt acts as an agent interacting with the LLMs, which is demonstrated in Figure~\ref{fig:all}. We uses the generated sequence of in-context examples and the user query to construct the final prompt when interacting with LLMs. The demonstration of the prompt format used in the prompt creator is shown in Appendix~\ref{adx:a_2}, while the exemplary demonstration of user query and the in-context example can be found in Figure~\ref{fig:all}. Intuitively, if the response of LLMs is correct, we update the policy network to increase the probability of generating the same sequence of the in-context samples. Conversely, we update the policy network to reduce the probability of generating that sequence.

\section{Experiment}
\subsection{Experimental Setup}
\subsubsection{Dataset} 
We evaluate our approach on two public datasets. The following datasets are chosen because they encompass a variety of text-to-text generation tasks and contain repetitive QA patterns.

\textbf{Alpaca}~\cite{taori2023stanford} dataset consists of 52,000 instruction-response pairs generated by OpenAI’s text-davinci-003 model. These instructions are expanded from existing open-source instruction datasets, covering a wide range of tasks from simple question-answering to complex reasoning and programming issues, thus providing strong cross-scenario adaptability.

\textbf{Dolly}~\cite{conover2023free} contains 15,000 human-annotated instruction-following records across tasks like brainstorming, classification, closed question-answering, text generation, and summarization. Annotators were instructed to avoid using online data, except Wikipedia for certain categories, and to refrain from using generative AI in crafting instructions or responses, ensuring the dataset's quality and authenticity.

\subsubsection{Baselines and Configurations}
We compare GRL-Prompt with three baselines: (a) Random selection: We randomly selected one example and two additional examples from the training set as the baselines in our experiments. (b) PromptPG~\cite{lu2022dynamic}, which utilizes policy gradient to learn to select in-context examples from a small amount of training data and then constructs the corresponding prompt for the test example. 
(c) CoT prompting~\cite{wei2022chain}, which is a widely validated approach of in-context learning. (d) Pillow~\cite{qi2023pillow}, which designs a matching network to select in-context examples and follows a reinforcement learning process to update its parameters. (e) UDR~\cite{li2023unified}, which proposes a bi-encoder architecture to represent candidate in-context examples and designs a ranking training algorithm to retrieve these examples.\\
\indent We randomly selected 1,800 data items: 200 for RL training, 800 for validation, and 800 for testing. Experiments for both the baselines and our GRL-Prompt were repeated three times, and the average metric is reported in Table~\ref{tab:results}. We implemented the experiments on a single NVIDIA A100 GPU for model training and evaluation.
\subsubsection{Evaluation Metric}
ROUGE is widely adopted to evaluate the informational completeness and coverage of generated texts by calculating the recall of generated text to reference text. We use ROUGE-1, ROUGE-2~\cite{lin2004rouge}, and ROUGE-L~\cite{lin2004automatic} as the evaluation metrics in this experiment to measure the quality of the generated text at different granularities. BLEU, on the other hand, measures the accuracy of the generated text by comparing the similarity between the generated text and the reference text~\cite{shang2021multimodal}. It places greater emphasis on precision matching than ROUGE.

\begin{figure*}[tb!]
\vspace{-8mm}
    \centering
    \scalebox{0.83}{
    \subfigure[
	ROUGE-1.] 
	{\includegraphics[width=1.65in,height=1.5in]{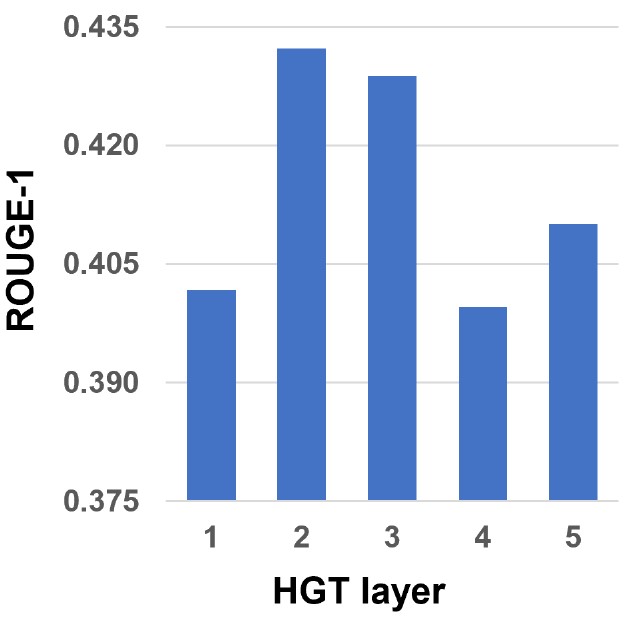}}
	\subfigure[
	ROUGE-2.] 
	{\includegraphics[width=1.65in,height=1.5in]{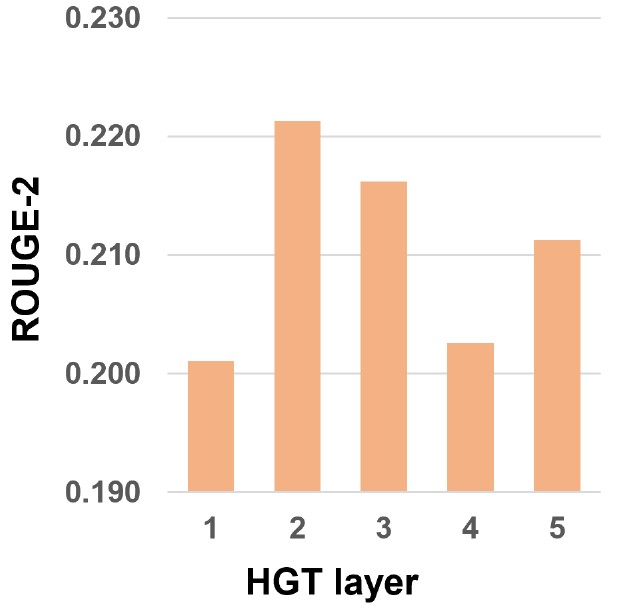}}
 \subfigure[\label{fig:8}
	ROUGE-L.] 
	{\includegraphics[width=1.65in,height=1.5in]{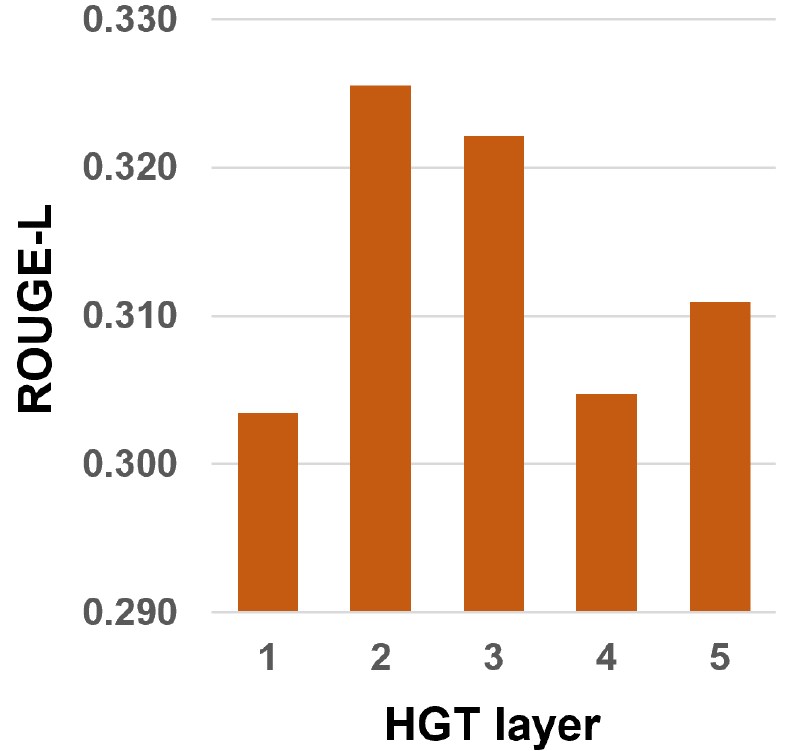}}
 \subfigure[\label{fig:9}
	BLEU.] 
	{\includegraphics[width=1.65in,height=1.5in]{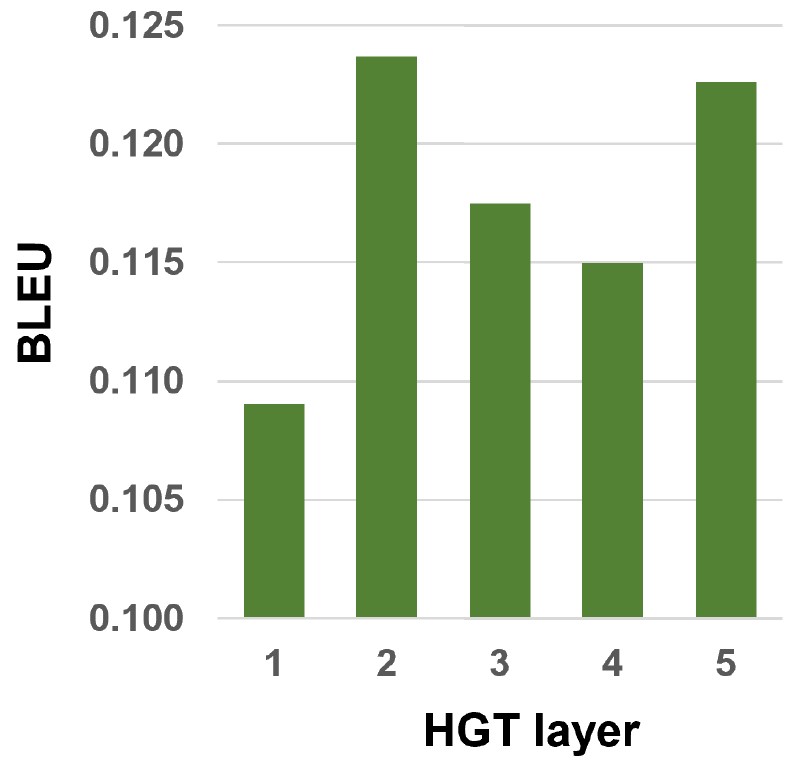}}
  }
 \caption{Sensitivity analysis on HGT layer.}\label{fig:6}
  \vspace{-4mm}
\end{figure*}
\begin{figure*}[tb!]
    \centering
    \scalebox{0.83}{
    \subfigure[\label{fig:10}
	ROUGE-1.] 
	{\includegraphics[width=1.65in,height=1.5in]{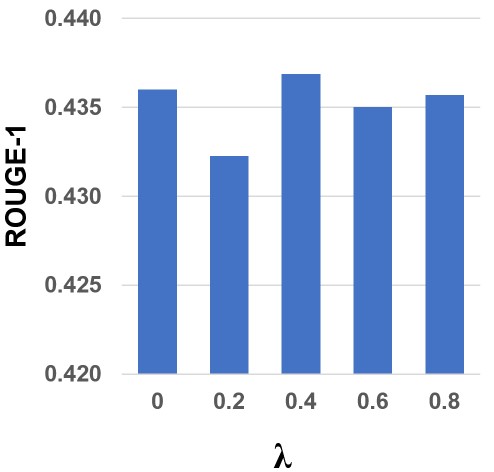}}
	\subfigure[\label{fig:11}
	ROUGE-2.] 
	{\includegraphics[width=1.65in,height=1.5in]{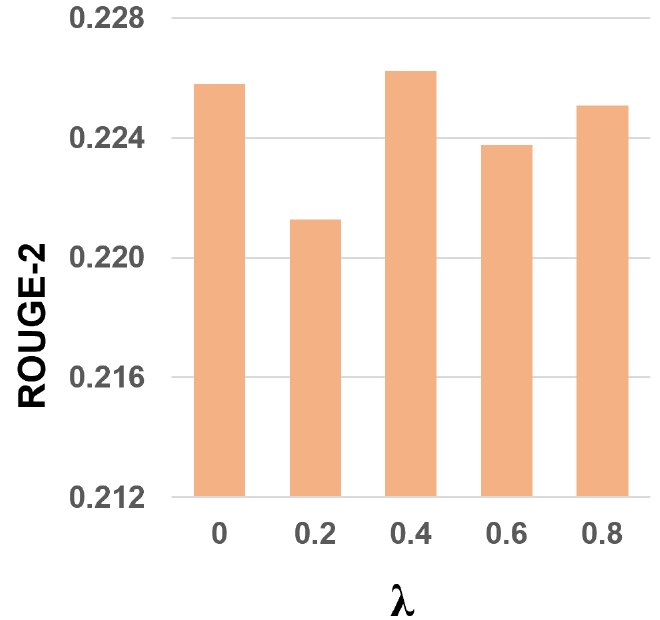}}
 \subfigure[\label{fig:12}
	ROUGE-L.] 
	{\includegraphics[width=1.65in,height=1.5in]{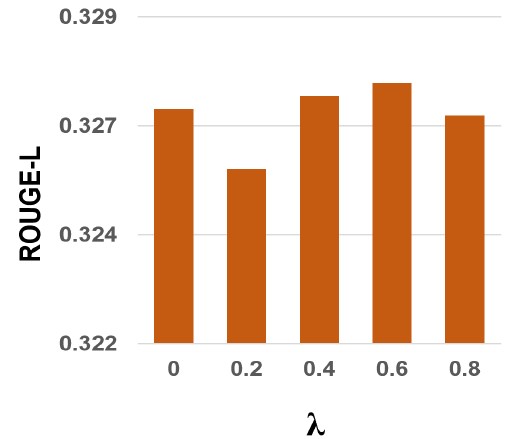}}
 \subfigure[\label{fig:13}
	BLEU.] 
	{\includegraphics[width=1.65in,height=1.5in]{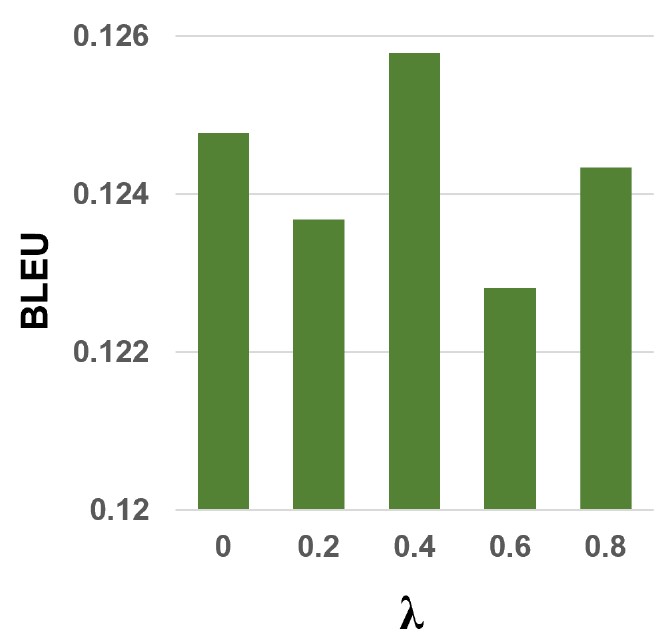}}
 }
 \caption{Sensitivity analysis on $\lambda$.}\label{fig:7}
 \vspace{-6mm}
\end{figure*}
\subsection{Performance Analysis }
As shown in Table~\ref{tab:results}, GRL-Prompt can enhance the in-context learning performance on both public datasets regarding Rouge and BLEU metrics compared with the above baselines.
Our method outperforms all baseline methods across two datasets, showing an average improvement of 0.10 on ROUGE-1, 0.07 on ROUGE-2, 0.07 on ROUGE-L, and 0.05 on BLEU. It indicates that GRL-Prompt learns the order-sensitive policy from the constructed knowledge graph to generate the optimal sequence of in-context examples for prompt optimization, effectively reducing the uncertainty and noise introduced by the random selection and permutation of the in-context examples. It is worth noting that CoT prompting performs worse than random selection because the examples of CoT are manually designed for the specific task.\\
\indent As shown in Figures~\ref{fig:2} and~\ref{fig:3}, we compare the trend of training loss with UDR, the baseline that performed the best on nearly all metrics. 
Models trained with the GRL-Prompt method (i.e., Ours-GPT-4, Ours-GPT-3, and Ours-LLaMA) demonstrate a significant reduction in loss, while converging faster in later epochs, reflecting strong learning performance and stability. In contrast, UDR method (i.e., UDR-GPT-4, UDR-GPT-3, and UDR-LLaMA) start with lower loss values, and the rate of decrease diminishes as training progresses, with fluctuations and consistently higher final loss values. Moreover, we conduct extensive experiments to compare the trend of training loss with all baselines, which can be found in Appendix~\ref{adx:a_3}. The fast convergence rate and lower loss observed with the GRL-Prompt method are primarily attributed to the designated embedding-based reward, which effectively incorporates feedback from LLMs during the prompt optimization phase. Additionally, the detailed case study can be found in Appendix~\ref{adx:a_4}.

\begin{table}[t!]
\centering
\scalebox{0.72}{
\begin{tabular}{l|llll}
\hline
\textbf{Method}      & \textbf{ROUGE-1} & \textbf{ROUGE-2} & \textbf{ROUGE-L} & \textbf{BLEU} \\ \hline
\textbf{Ours} &0.44   &0.23   &0.34  &0.15    \\ \hline
\textbf{-w/o KG}    &0.38($\downarrow$0.06)  &0.18($\downarrow$0.05)  &0.30($\downarrow$0.04)  &0.12($\downarrow$0.03) \\ \hline
\textbf{-w/o RF}    &0.37($\downarrow$0.07) &0.16($\downarrow$0.07) &0.28($\downarrow$0.06) &0.10($\downarrow$0.05)  \\  \hline
\end{tabular}
}
\caption{Performances of GRL-Prompt in ablation study on Alpaca dataset using GPT-4}
\vspace{-5mm}
\label{tab:ablation}
\end{table}

\subsection{Ablation Study}
To validate the effectiveness of different components in GRL-Prompt, we design two variants of GRL-Prompt and compare the performance of these variants with GRL-Prompt in this section. The results of comparison are shown in Table~\ref{tab:ablation}. GRL-Prompt w/o knowledge graph (KG) excludes the processes of KG construction and learning, and directly represents the user query and the candidate examples using the embedding from a pre-trained language model. The results show that GRL-Prompt outperforms over GRL-Prompt w/o KG by margins of 0.03-0.06 on four different evaluation metrics, indicating the correlation of the user query and the candidate examples encoded by KG enhances the performance of GRL-Prompt. GRL-Prompt w/o reward feedback (RF) replaces the RL-based prompt optimization with k-NN method for selecting the in-context examples, setting the number of examples to form the prompt at 2~\cite{nie2022improving}. The variant results in drops of 0.05 to 0.07 across four evaluation metrics, inferring the necessity of reward feedback for prompt optimization. Additionally, the overall experimental results on two datasets using three LLMs can be found in Appendix~\ref{adx:a_5}.
\vspace{-2mm}
\subsection{Sensitivity Analysis}
We study the impact of key parameters in GRL-Prompt, including the HGT layer and the $\lambda$ in the reward on Dolly dataset, the results are illustrated in Figure~\ref{fig:6} and~\ref{fig:7}, respectively. 

\begin{enumerate}[label=\arabic*)]
  \vspace{-2mm}
  \item \textit{Impact of the total number of layer in the HGT:} As shown in Figure~\ref{fig:6}, increasing the number of layers generally enhances our model's performance, while start dropping after a certain point. It is observed that our model achieves the best performance when the layer is set to be 2. Intuitively, continuously increasing the number of layers introduces the over-smoothing problem in graph learning, resulting in an unexpected decline in performance. 
  \vspace{-8mm}
  \item \textit{Impact of $\lambda$ in the reward:} As observed in Figure~\ref{fig:7}, the performance of the model achieves the best when $\lambda= 0.4$, showing an optimal trade-off point between the fuzzy textual similarity and the cosine embedding similarity in most cases. It also indicates that over-reliance on fuzzy textual similarity $\mathcal{R}_{m}$ aggravates the performance of GRL-Prompt. However, using the embedding-based reward that incorporates two components of rewards can help improve the performance of GRL-Prompt.
\end{enumerate}
\vspace{-5mm}




\section{Conclusion}
\vspace{-2mm}
In this paper, we propose a LLMs-agnostic prompt optimization framework called GRL-Prompt. This framework first constructs a knowledge graph from the user instruction and the candidate examples, learning with HGT to better encode their correlation. Afterwards, the policy network is designed to find the optimal sequence of in-context examples by incorporating the PEC and ICMN to classify the order of examples and select the relevant ones. We also conduct comprehensive experiments on two widely-used public datasets and compare the proposed GRL-Prompt with four baseline methods. The results and analysis reveal that the proposed framework achieves state-of-the-art performance across different LLMs in the in-context learning paradigm.

\newpage

\bibliography{main}

\appendix
\clearpage
\section{Appendix}

\renewcommand{\thetable}{1}
\begin{table*}[]
\vspace{2mm}
\centering
\scalebox{0.73}{
\begin{tabular}{llcccc|cccc}
\hline
\multirow{2}{*}{Model} & \multirow{2}{*}{Method} & \multicolumn{4}{c|}{Alpaca}        & \multicolumn{4}{c}{Dolly}          \\ \cline{3-10} 
                       &                         & Rouge-1 & Rouge-2 & Rouge-L & BLEU & Rouge-1 & Rouge-2 & Rouge-L & BLEU \\ \hline
\multirow{3}{*}{GPT-4} & \textbf{Ours}     & 0.44    & 0.23    & 0.34    & 0.15 & 0.43    & 0.21    & 0.32    & 0.11 \\ \cline{2-10} 
                       & -w/o KG                 & 0.38($\downarrow 0.06$)    & 0.18($\downarrow 0.05$)    & 0.30($\downarrow 0.04$)    & 0.12($\downarrow 0.03$) & 0.41($\downarrow 0.02$)    & 0.20($\downarrow 0.01$)    & 0.31($\downarrow 0.01$)    & 0.08($\downarrow 0.03$) \\
                       & -w/o RF                 & 0.37($\downarrow 0.07$)    & 0.16($\downarrow 0.07$)    & 0.28($\downarrow 0.06$)    & 0.10($\downarrow 0.05$) & 0.37($\downarrow 0.06$)    & 0.17($\downarrow 0.04$)    & 0.27($\downarrow 0.05$)    & 0.08($\downarrow 0.03$) \\ \hline
\multirow{3}{*}{GPT-3} & \textbf{Ours}     & 0.50    & 0.27    & 0.39    & 0.19 & 0.47    & 0.26    & 0.37    & 0.15 \\ \cline{2-10} 
                       & -w/o KG                 & 0.46($\downarrow 0.04$)    & 0.23($\downarrow 0.04$)    & 0.37($\downarrow 0.02$)    & 0.16($\downarrow 0.03$) & 0.44($\downarrow 0.03$)    & 0.24($\downarrow 0.02$)    & 0.35($\downarrow 0.02$)    & 0.14($\downarrow 0.01$) \\
                       & -w/o RF                 & 0.41($\downarrow 0.09$)    & 0.19($\downarrow 0.08$)    & 0.32($\downarrow 0.07$)    & 0.12($\downarrow 0.07$) & 0.40($\downarrow 0.07$)    & 0.20($\downarrow 0.06$)    & 0.32($\downarrow 0.05$)    & 0.11($\downarrow 0.04$) \\ \hline
\multirow{3}{*}{LLaMA} & \textbf{Ours}     & 0.45    & 0.23    & 0.35    & 0.15 & 0.41    & 0.21    & 0.31    & 0.12 \\ \cline{2-10} 
                       & -w/o KG                 & 0.35($\downarrow 0.10$)    & 0.14($\downarrow 0.09$)    & 0.26($\downarrow 0.09$)    & 0.08($\downarrow 0.07$) & 0.39($\downarrow 0.02$)    & 0.18($\downarrow 0.03$)    & 0.30($\downarrow 0.01$)    & 0.11($\downarrow 0.01$) \\
                       & -w/o RF                 & 0.32($\downarrow 0.13$)    & 0.12($\downarrow 0.11$)    & 0.24($\downarrow 0.11$)    & 0.07($\downarrow 0.08$) & 0.36($\downarrow 0.05$)    & 0.15($\downarrow 0.06$)    & 0.28($\downarrow 0.05$)    & 0.09($\downarrow 0.02$) \\ \hline
\end{tabular}
}
\vspace{3mm}
\caption{Performances of GRL-Prompt in supplementary ablation study.}
\label{adx:tab}
\end{table*}
\subsection{The Details of Heterogeneous Graph Transformer (HGT)}~\label{adx:a_1}
The HGT can model heterogeneous relations in our knowledge graph using a multi-head attention mechanism. It calculates the multi-head attention between a given node and its neighbor nodes with different relations, and then uses the normalized attention as the weight for the neighbor nodes to update the embedding of the given node. Taking the node $v_{i}$ as an example, the embedding learning process can be formulated as:

\renewcommand{\theequation}{1}
\begin{equation}
\hspace{-4mm}
\begin{split}
    &\emph{Q}^{h}\left(v_{j}\right)= \emph{X}^{l-1}\left(v_{j}\right)\cdot Q\text{-}Lin^{h}_{\phi(v_{j})} \\
    &\emph{K}^{h}\left(v_{i}\right) = \emph{X}^{l-1}\left(v_{i}\right)\cdot K\text{-}Lin^{h}_{\phi (v_i)}\\
    &att^{h}\!\!\left(v_i\!,\!r_{ij}\!,\!v_j\!\right)\!\!=\!\!(\! \emph{Q}^{h}\left(v_{j}\right)\!\cdot\!\Theta^{att}_{r_{ij}}\!\cdot\!\emph{K}^{h}\!\left(v_{i}\right)\!^{T}\!)\!\cdot\!\frac{\mu_{\langle\phi(\!v_i\!),r_{ij},\phi(\!v_j\!)\rangle}}{\sqrt{d}}
\end{split}
\end{equation}
We use $l\in[1,...,L]$ to denote the $l$-th HGT layer and $L$ is the total number of layers. Given a target node $v_j$, and all its neighbors $v_{i}\in\mathcal{N}(v_j)$, we first project the target node into the $h$-th query vector $\emph{Q}^{h}\left(v_{j}\right)$ with a type-specific linear transformation matrix $Q\text{-}Lin^{h}\in\mathcal{R}^{d\times\frac{d}{H}}$, where $H$ is the number of attention heads. $\phi(\cdot)$ is the type mapping function. We also project the neighbor node $v_{i}$ into the $h$-th key vector $\emph{K}^{h}\left(v_{i}\right)$ on the same dimension. Next, we apply a relation-specific weight matrix $\Theta^{att}_{r_{ij}}\in\mathcal{R}^{\frac{d}{H}\times\frac{d}{H}}$ to obtain the $h$-th attention head $att^{h}\left(v_i, r_{ij}, v_j\right)$ and $\sqrt{d}$ acts a scaling factor. Moreover, since not all the relations contribute equally to the target nodes, a learnable vector $\mu$ associated with the relation triplet $\langle\phi(v_i), r_{ij}, \phi(v_j)\rangle$ acts as a scaling factor for this triplet relationship. Finally, the weight of the neighbor node $w^{h}\left(v_{i}\right)$ is calculated by \textit{softmax} normalization. The attentive aggregation of different heads across neighbor nodes with various relations for updating node embedding of $v_{j}$ is defined as:
\renewcommand{\theequation}{2}
\begin{equation}
\begin{split}    
    \emph{X}^{l}(v_j)\!=\!\frac{1}{|\mathcal{N}(v_j)|}\!\sum_{v_i\in\mathcal{N}(v_j)}\!\!\!\!MLP(\!\!\!\!\concat_{h\in\left[1,H\right]}\!\!\!\!w^{h}(v_{i})\!\cdot\!\emph{K}^{h}\left(v_{i}\right)) 
\end{split}
\end{equation}
where $\concat$ represents concatenation. We first aggregate all embedding of neighbor nodes, and then concatenate all $H$ heads. After that, the node embedding of $v_j$ is updated by a shallow multi-layer perceptron (MLP).

\subsection{Prompt Template}~\label{adx:a_2}
\justifying
The prompt template is designed to interact with different LLMs using the proposed GRL-Prompt. Figure~\ref{fig:a5} demonstrates the designated prompt template format. Notably, the sequence of the in-context examples $\vec{\emph{P}}_{ic} = [p_{ic}^{1}, ..., p_{ic}^{K}]$ in the prompt template is generated by GRL-Prompt, and the user query $q$ is derived from the test data. In Figure~\ref{fig:a5}, the placeholders (marked as red) denote the in-context examples and the user query.
\renewcommand{\thefigure}{1}
\begin{figure}[h]
    \centering
    \includegraphics[width=1.05\linewidth]{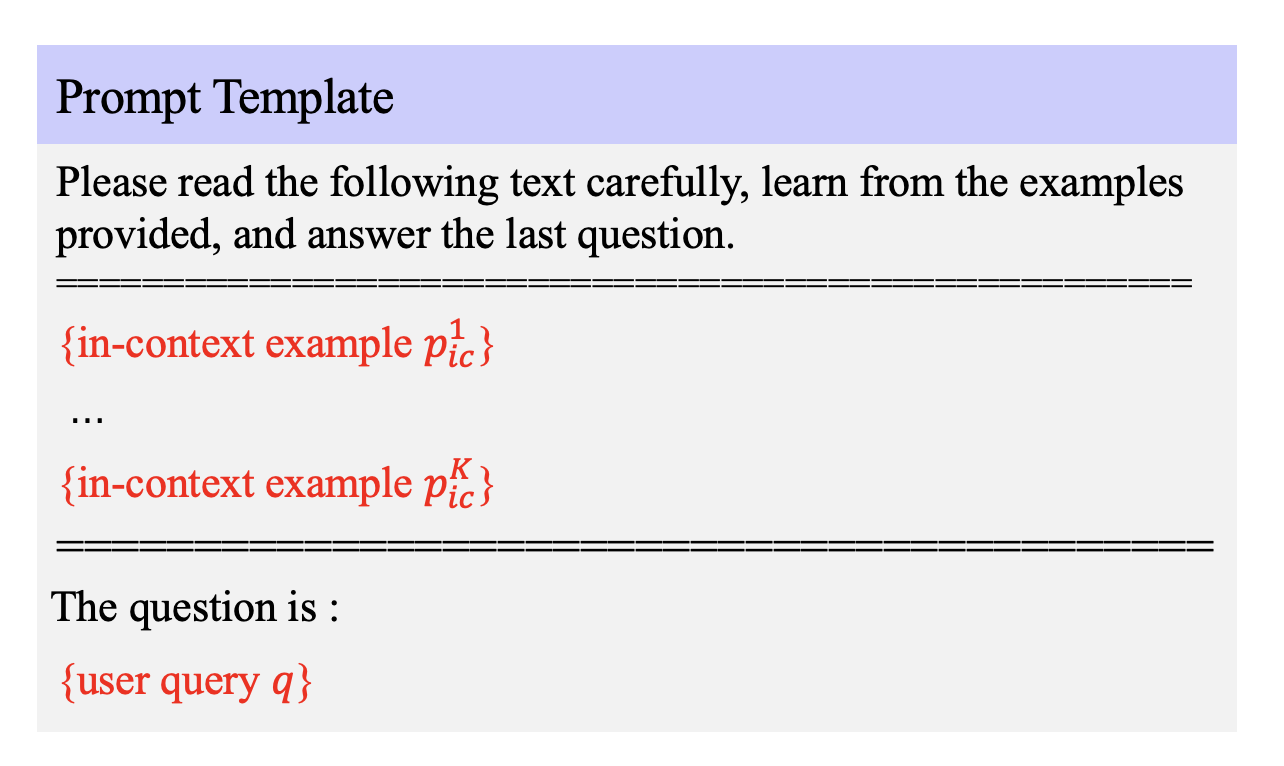}
  \caption{The demonstration of the prompt template}
  \label{fig:a5}
\hspace{-3mm}
\end{figure}

\subsection{Supplementary Comparison of The Training Loss Trend with Baselines.}~\label{adx:a_3}
We conduct extensive experiments to compare the trend of training loss with all baselines, as illustrated in the Figure~\ref{fig:appendix_loss}. The fast convergence rate and lower loss observed with the GRL-Prompt method are primarily attributed to the designated embedding-based reward, which effectively incorporates feedback from LLMs during the prompt optimization phase.
\renewcommand{\thefigure}{2}
\begin{figure*}[ht!]
    \scalebox{1.03}{ 
    \subfigure[GPT-4 on Alpaca] 
    {\includegraphics[width=2.0in,height=1.8in]{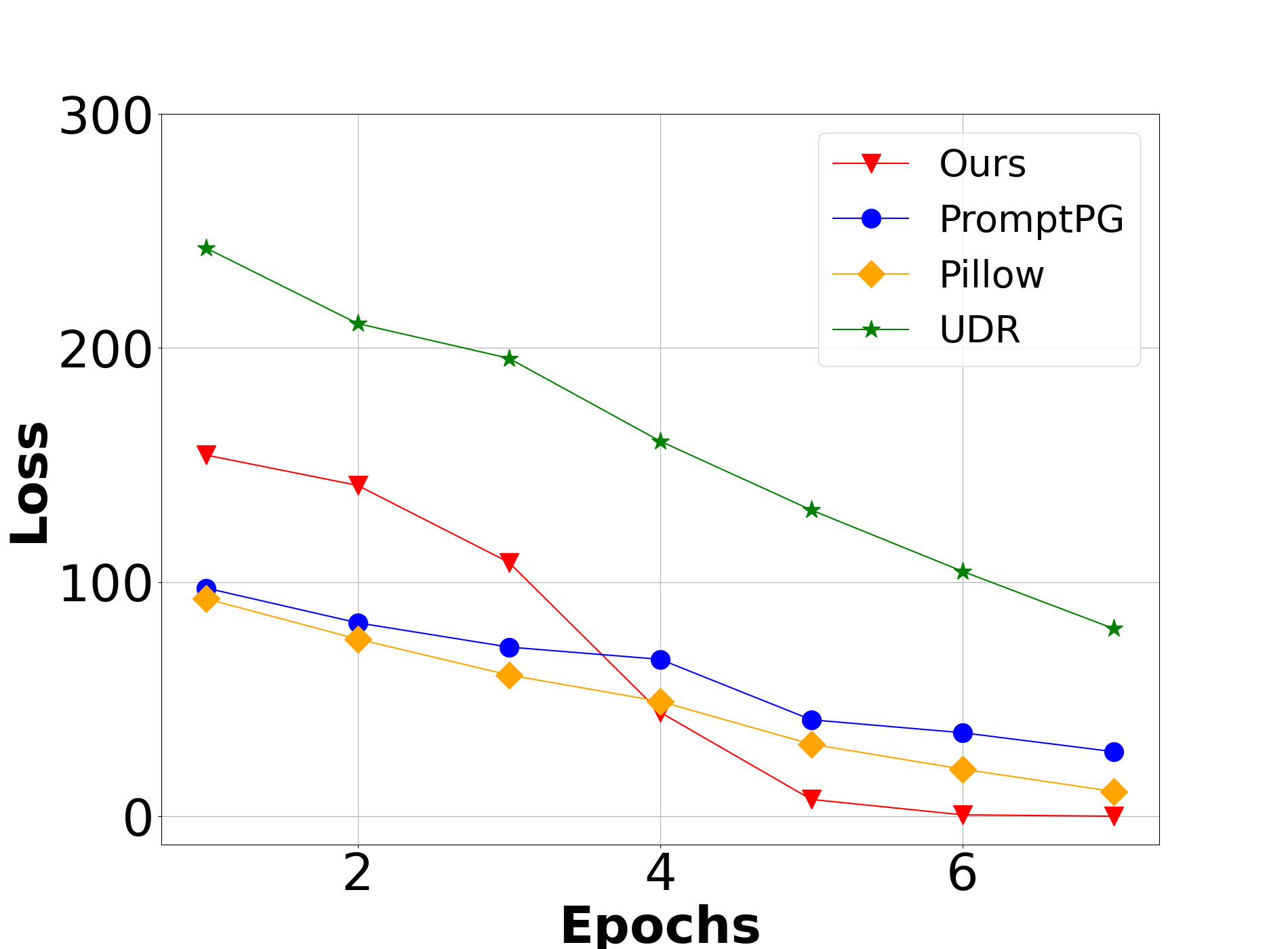}}
    \subfigure[GPT-3 on Alpaca] 
    {\includegraphics[width=2.0in,height=1.8in]{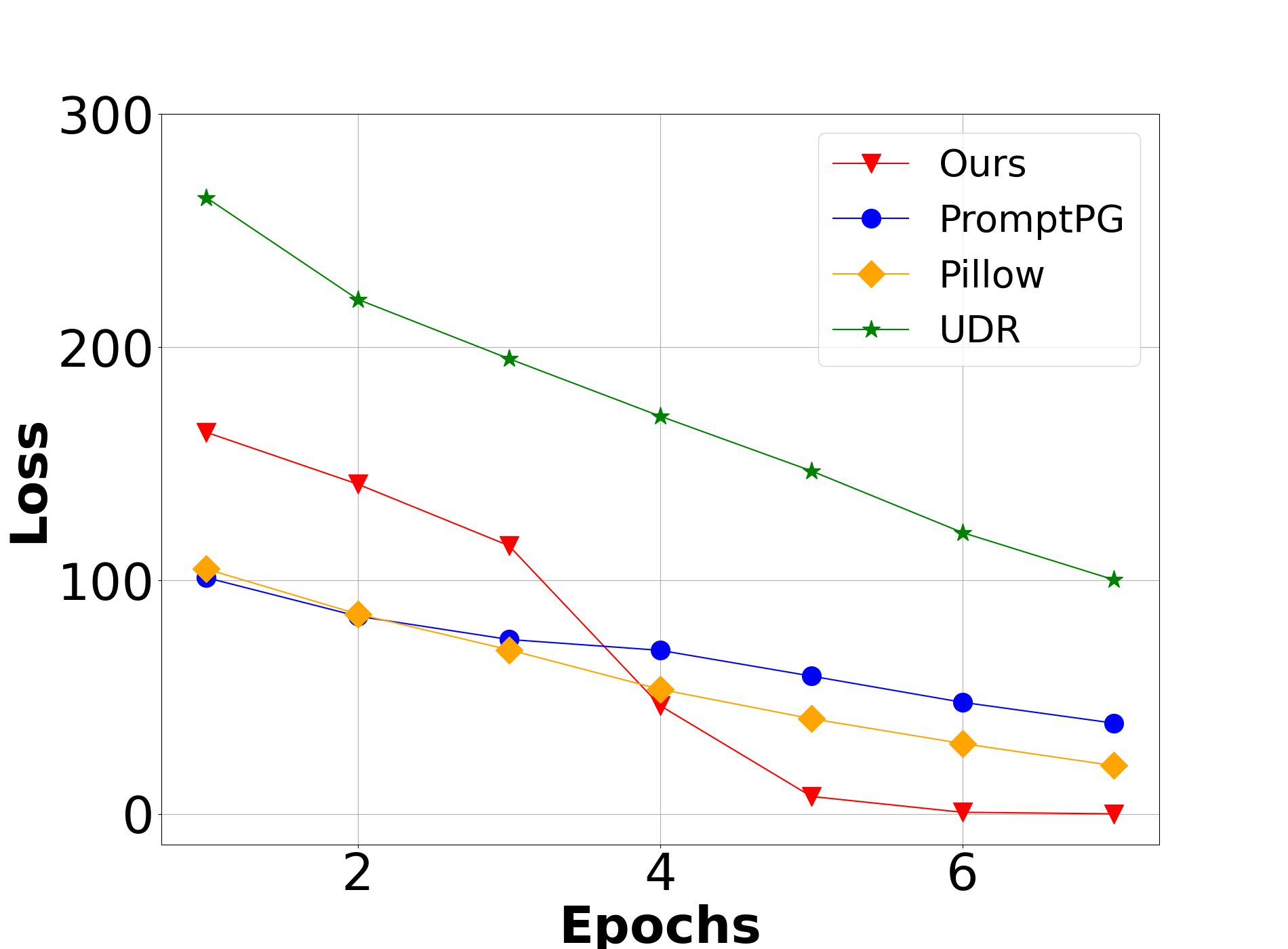}}
    \subfigure[LLaMA on Alpaca] 
    {\includegraphics[width=2.0in,height=1.8in]{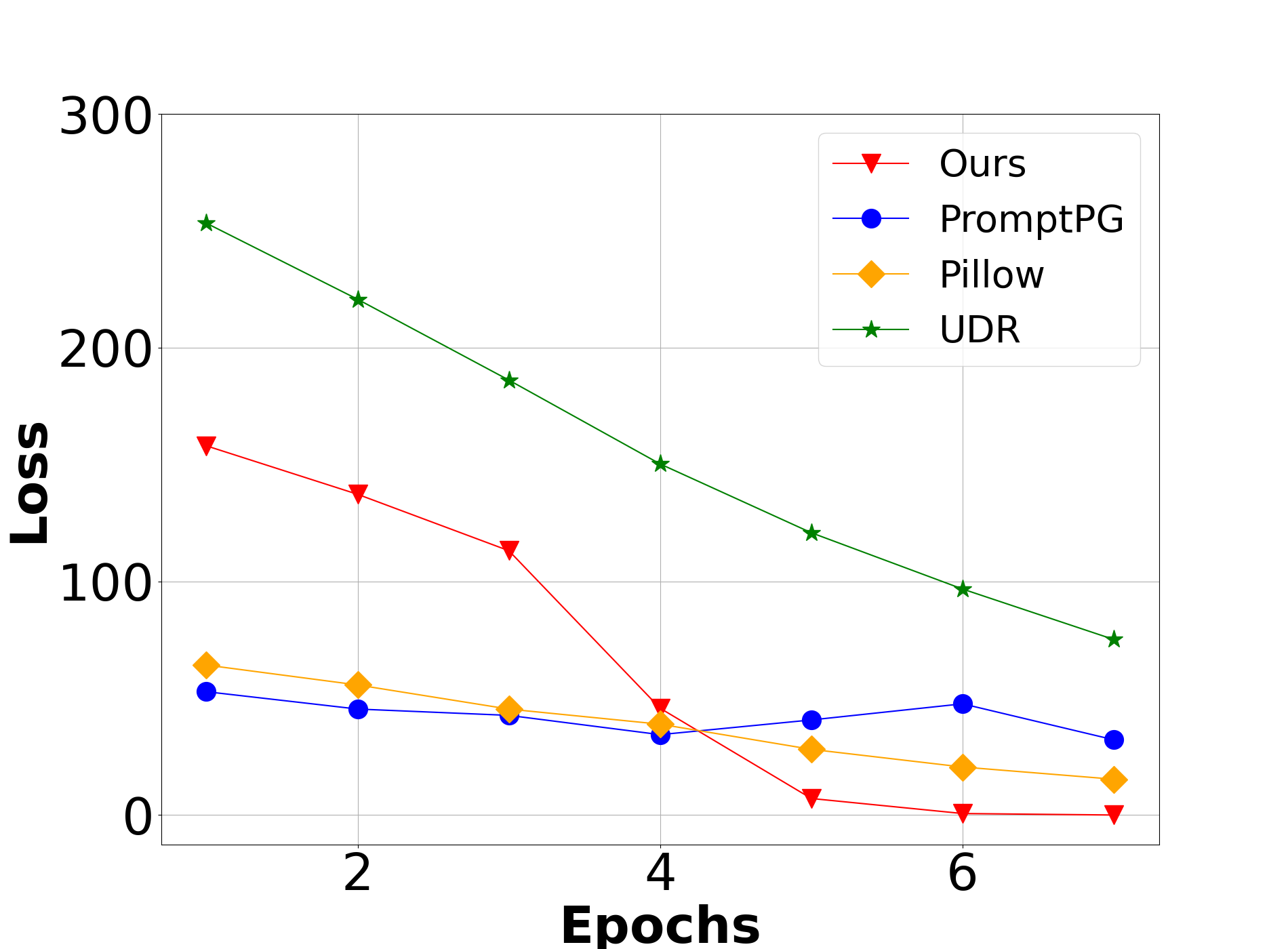}}
    }
    \scalebox{1.03}{
    \subfigure[GPT-4 on Dolly] 
    {\includegraphics[width=2.0in,height=1.8in]{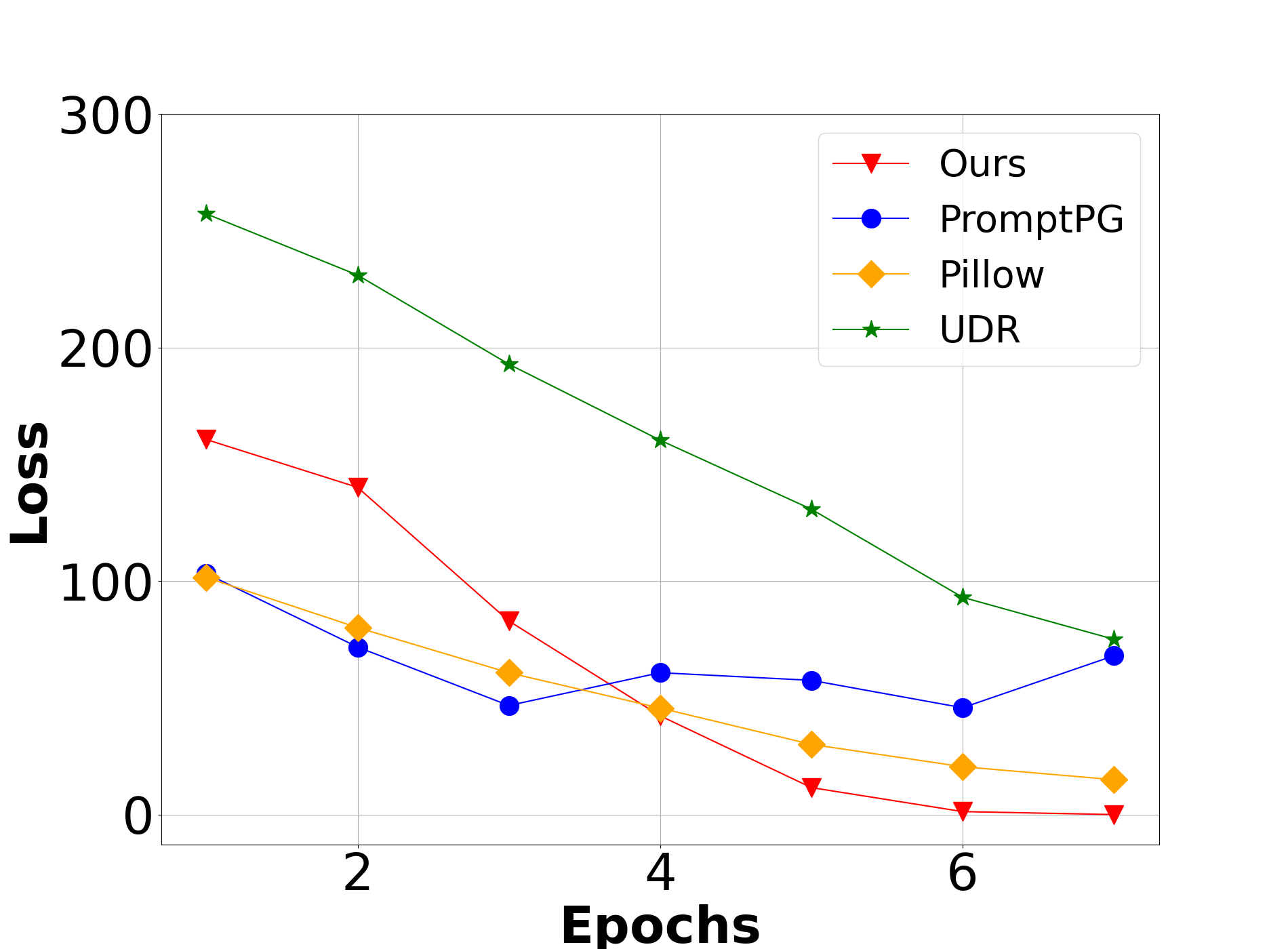}}
    \subfigure[GPT-3 on Dolly] 
    {\includegraphics[width=2.0in,height=1.8in]{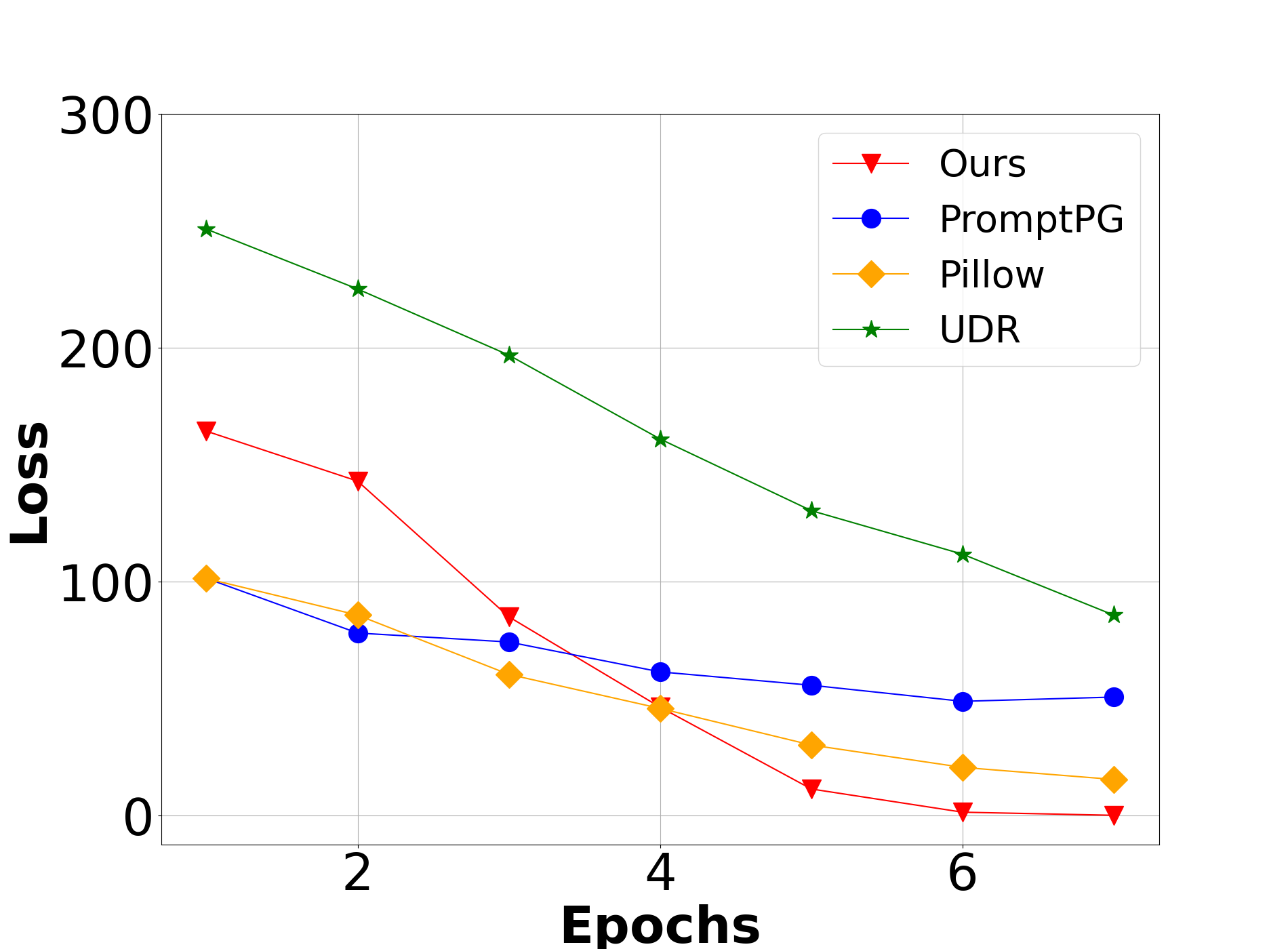}}
    \subfigure[LLaMA on Dolly] 
    {\includegraphics[width=2.0in,height=1.8in]{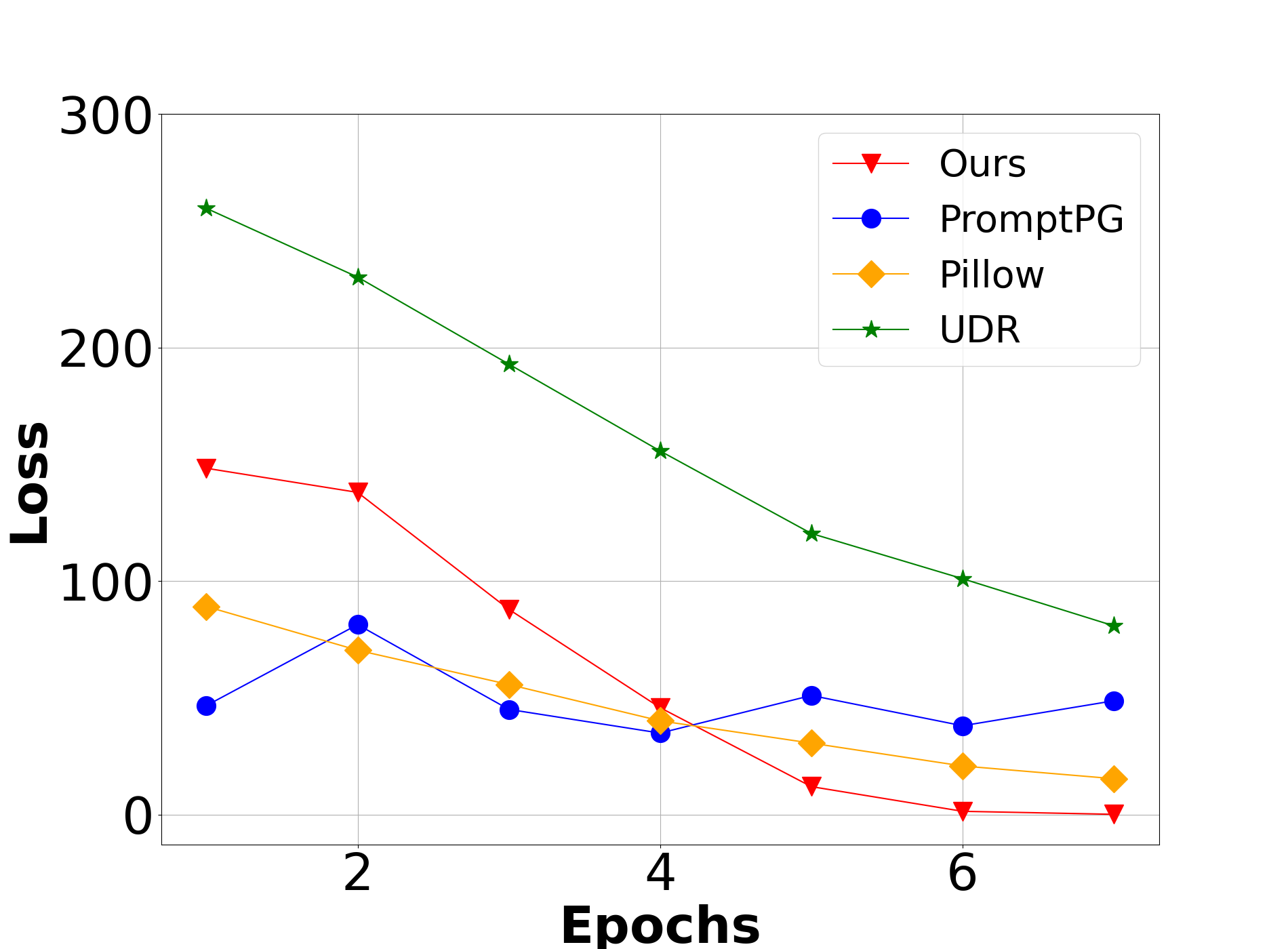}}
    }
    \vspace{2mm}
    \caption{The trend of training loss across different LLMs on two datasets.}\label{fig:appendix_loss}
\end{figure*}

\subsection{Case Study}~\label{adx:a_4}
Figures~\ref{fig:c1} to \ref{fig:c6} show the comparative case study of GRL-Prompt and other baseline methods, i.e.,  Random-1, Random-2, CoT, and PromptPG, Pillow, UDR respectively. It is evident that prompts optimized with GRL-Prompt result in more reasonable responses from LLMs compared to other baseline methods. For example, in Figure~\ref{fig:c1}, the generated prompt from GRL-Prompt (marked in blue) provides more instructive guidance for LLMs. Random-1, on the other hand, clearly gives an oversimplified prompt, leading to an undesired response from the LLMs. Similar results can also be found in Random-2, CoT, and PromptPG, Pillow, UDR (from Figures~\ref{fig:c2} to \ref{fig:c6}).


\renewcommand{\thefigure}{3}
\begin{figure*}[ht!]
    \centering
    \includegraphics[width=1\linewidth]{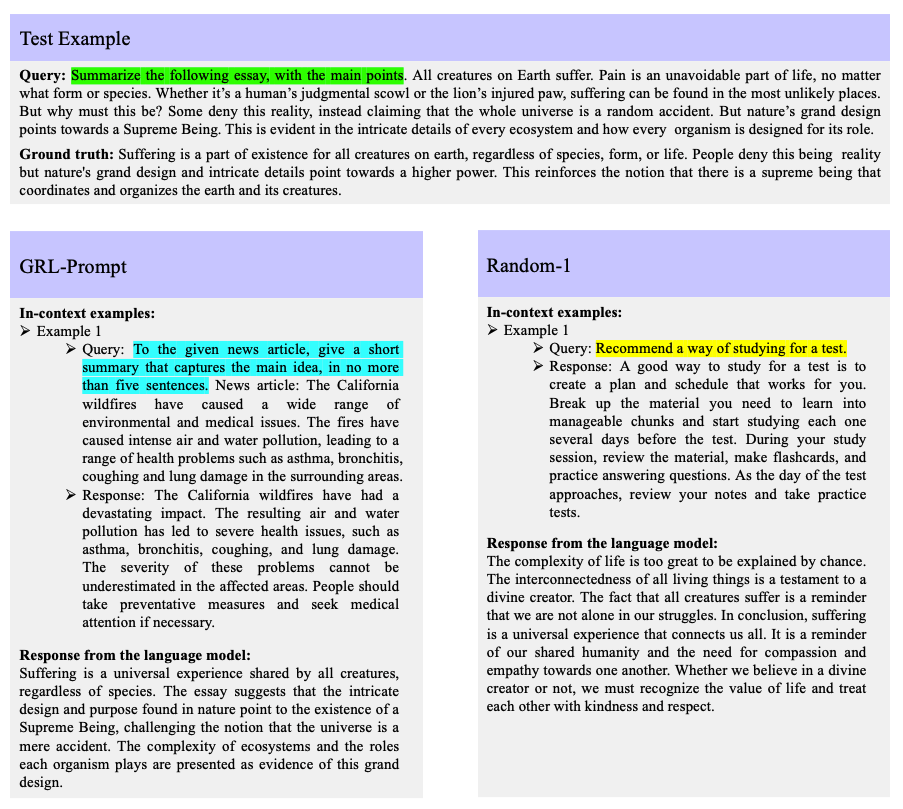}
  \caption{Comparative case study of GRL-Prompt and Random-1. The text highlighted in green indicates the keywords from the test example. In contrast, the text marked in blue represents the keywords in the in-context examples selected by GRL-Prompt, while the text highlighted in yellow corresponds to the keywords from the in-context examples chosen by Random-1.}
  \label{fig:c1} 
  
\end{figure*} 

\renewcommand{\thefigure}{4}
\begin{figure*}[!h]
    \centering
    \includegraphics[width=1\linewidth]{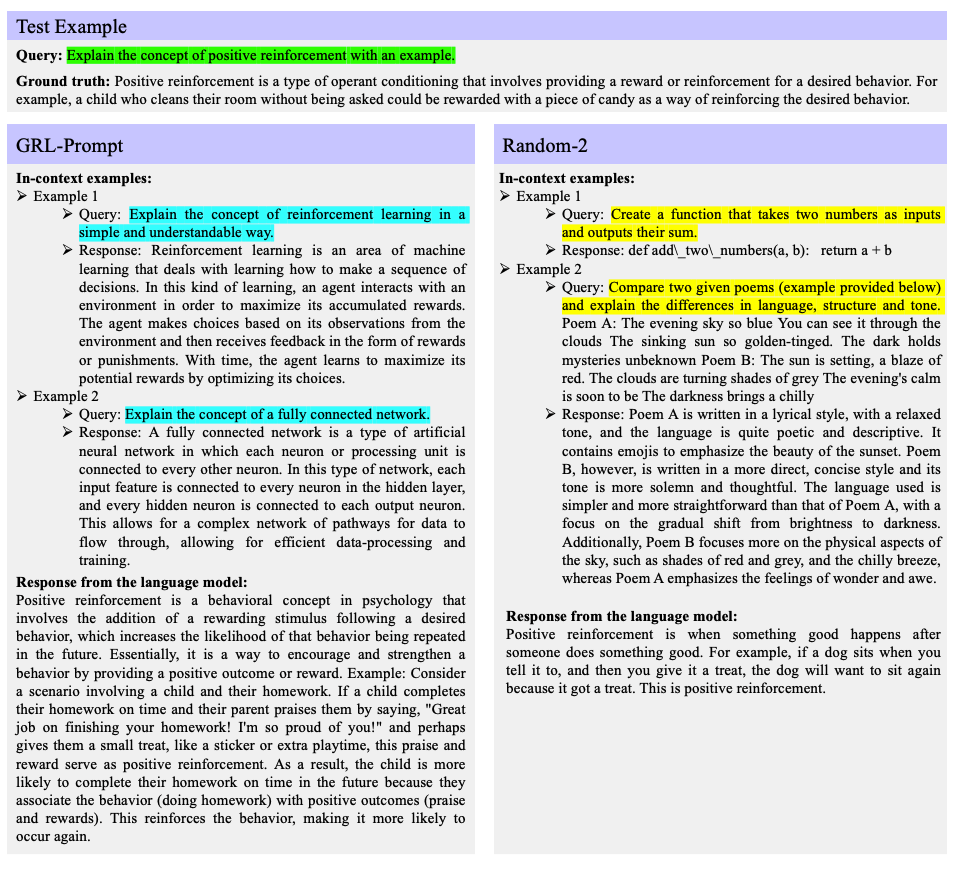}
  \caption{Comparative case study of GRL-Prompt and Random-2. The text highlighted in green indicates the keywords from the test example. In contrast, the text marked in blue represents the keywords in the in-context examples selected by GRL-Prompt, while the text highlighted in yellow corresponds to the keywords from the in-context examples chosen by Random-2.}
  \label{fig:c2} 
\end{figure*}

\renewcommand{\thefigure}{5}
\begin{figure*}[!h]
    \centering
    \includegraphics[width=1\linewidth]{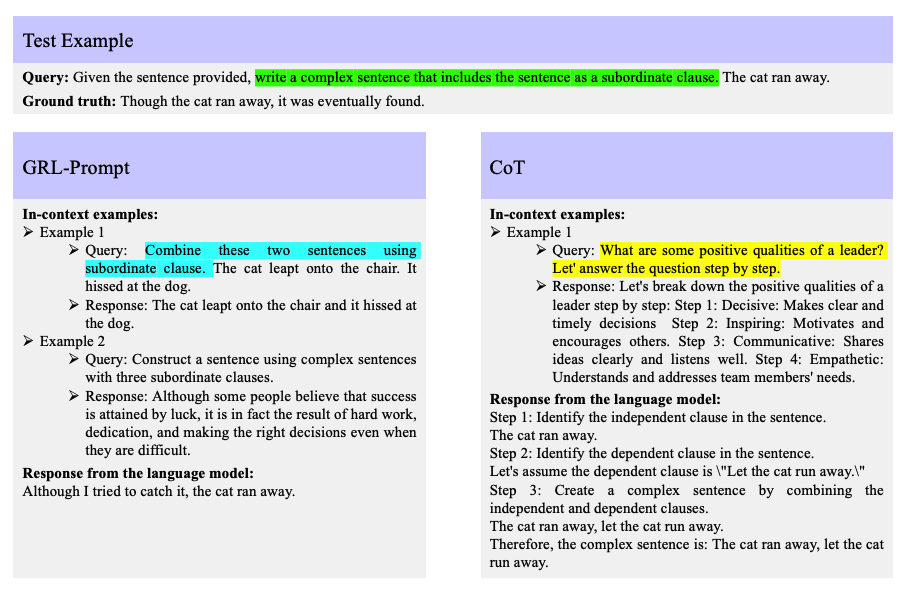}
  \caption{Comparative case study of GRL-Prompt and CoT. The text highlighted in green indicates the keywords from the test example. In contrast, the text marked in blue represents the keywords in the in-context examples selected by GRL-Prompt, while the text highlighted in yellow corresponds to the keywords from the in-context examples chosen by CoT.}
  \label{fig:c3} 
\end{figure*}

\renewcommand{\thefigure}{6}
\begin{figure*}[!h]
    \centering
    \includegraphics[width=1\linewidth]{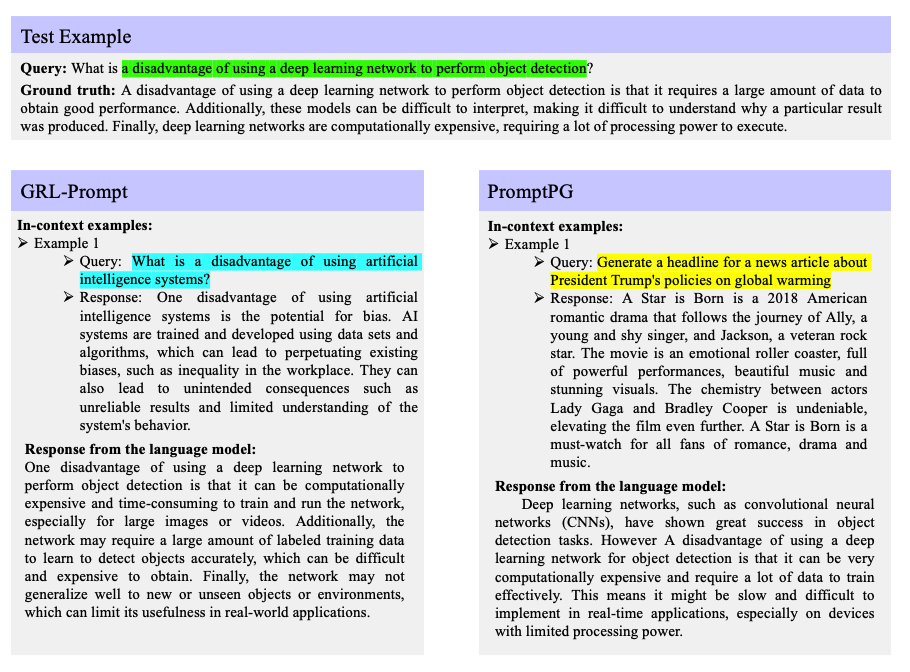}
  \caption{Comparative case study of GRL-Prompt and PromptPG. The text highlighted in green indicates the keywords from the test example. In contrast, the text marked in blue represents the keywords in the in-context examples selected by GRL-Prompt, while the text highlighted in yellow corresponds to the keywords from the in-context examples chosen by PromptPG.}
  \label{fig:c4} 
\end{figure*} 

\renewcommand{\thefigure}{7}
\begin{figure*}[!h]
    \centering
    \includegraphics[width=1\linewidth]{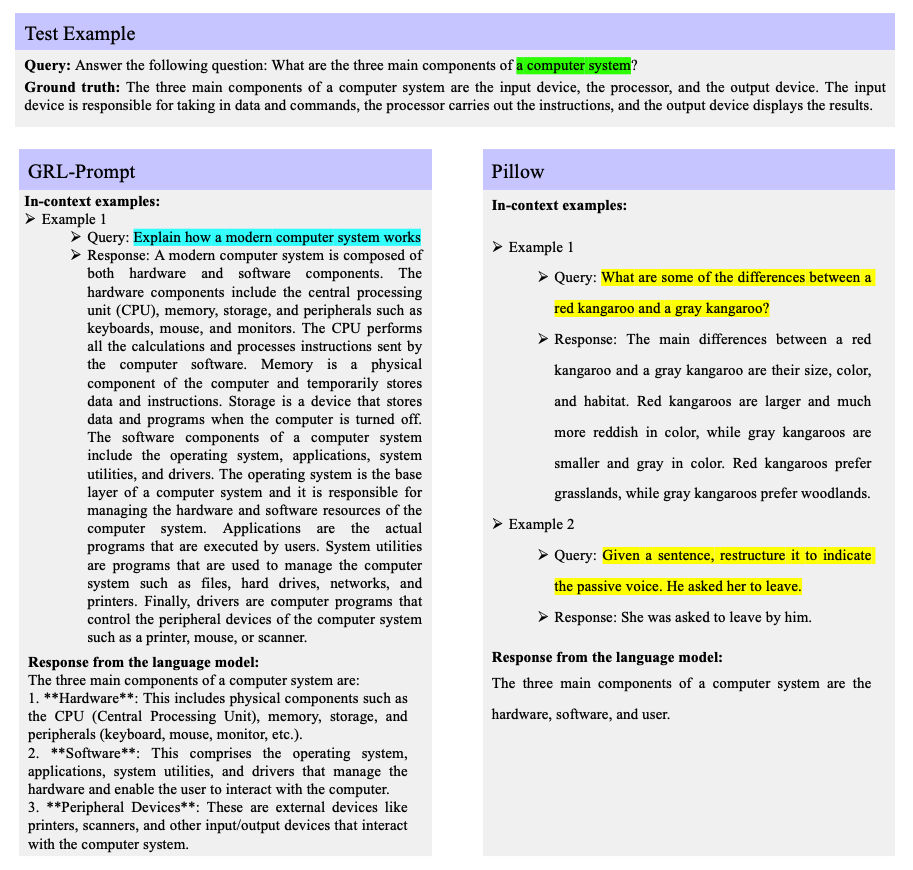}
  \caption{Comparative case study of GRL-Prompt and Pillow. The text highlighted in green indicates the keywords from the test example. In contrast, the text marked in blue represents the keywords in the in-context examples selected by GRL-Prompt, while the text highlighted in yellow corresponds to the keywords from the in-context examples chosen by Pillow.}
  \label{fig:c5} 
\end{figure*} 

\renewcommand{\thefigure}{8}
\begin{figure*}[!h]
    \centering
    \includegraphics[width=1\linewidth]{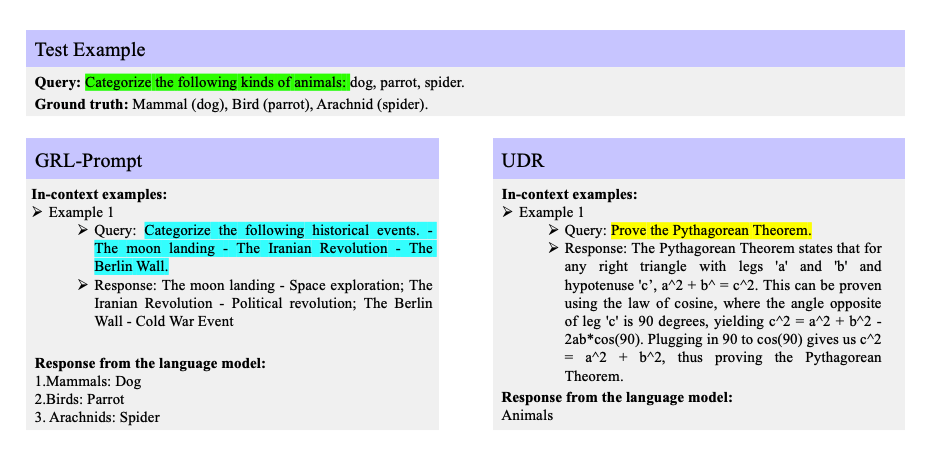}
  \caption{Comparative case study of GRL-Prompt and UDR. The text highlighted in green indicates the keywords from the test example. In contrast, the text marked in blue represents the keywords in the in-context examples selected by GRL-Prompt, while the text highlighted in yellow corresponds to the keywords from the in-context examples chosen by UDR.}
  \label{fig:c6} 
\end{figure*} 

\subsection{Supplementary Ablation Study} ~\label{adx:a_5}
\justifying
To further validate the effectiveness of different components in GRL-Prompt, we compare the two variants of GRL-Prompt with GRL-Prompt across different LLMs on two datasets. The overall results of the comparison are shown in Table~\ref{adx:tab}. The results show that GRL-Prompt outperforms both GRL-Prompt w/o KG and GRL-Prompt w/o RF across all different LLMs on two datasets, demonstrating an average improvement of 0.04 and 0.07, respectively. It highlights the significant contributions of two components in GRL-Prompt that enhance the performance of LLMs.

\subsection{Limitations}
The reward design in GRL-Prompt is based on automatic evaluation scores, which may not be sufficient for providing human feedback to the RL agent. Such scores focus on the contextual discrepancy between the generated and expected responses, which may not align with the results of human evaluations. In the future, we plan to integrate human feedback into the GRL-Prompt framework and investigate whether incorporating context-aware results and feedback can help enhance LLMs performance on different tasks.

\end{document}